\documentclass[12pt,letterpaper]{article}
\usepackage[lmargin=1.2in,rmargin=1.2in,tmargin=1in,bmargin=1in]{geometry}
\usepackage[bf,tiny]{titlesec}
\usepackage[utf8]{inputenc}
\usepackage[T1]{fontenc}

\usepackage[tt=false]{libertine}

\usepackage{microtype}

\usepackage{adjustbox}

\usepackage{todonotes}
\usepackage{longtable}
\usepackage{hyperref}

\usepackage{setspace}
\usepackage{gb4e}
\usepackage[sort]{natbib}
\bibpunct[: ]{(}{)}{,}{a}{}{,}
\usepackage{amsmath}
\usepackage{amssymb}
\usepackage{amsfonts}
\usepackage{graphicx}
\usepackage{svg} 
\usepackage{subcaption}
\usepackage{tikz}
\usepackage{adjustbox}
\usepackage{xcolor} 
\usepackage{fullpage}

\usepackage{authblk}

\title{Semantics drives analogical change in Germanic strong verb paradigms: a phylogenetic study}
 \date{February 24, 2025}

\author{Alexandru Craevschi}
\author{Sarah Babinski}
\author{Chundra Cathcart}

\affil{Institute for the Interdisciplinary Study of Language Evolution\\ University of Zurich}

\begin{document}

\maketitle


\begin{abstract}
    \noindent 
    A large body of research on morphological paradigms makes the prediction that irregular morphological patterns of allomorphy are more likely to emerge and persist when they serve to mark important functional distinctions. 
    More specifically, it has been observed that in some Germanic languages in which narrative past tense is expressed by the past participle, there is a greater affinity for stem allomorphy shared by preterite forms and past participles to the exclusion of present forms (the so-called ABB pattern), as it serves to enhance marking of the binary semantic opposition between present and past. Using data from 107 cognate verbs attested across 14 archaic and contemporary Germanic languages and a novel hierarchical phylogenetic model, we show that there is a greater long-term preference for this alternation pattern in situations where narrative past tense has been extended to the past participle, confirming this hypothesis. 
    We further elucidate the mechanisms underlying this association, demonstrating that this association holds because verbs with the ABB pattern are more likely to preserve it in situations where it marks an important binary semantic opposition; however, there is less evidence that the ABB pattern is extended to verbs with different patterns under the same circumstances. 
    These results bear on debate as to whether the distribution of irregularity we observe cross-linguistically is due primarily to (1) the preservation of irregular patterns or (2) an active drive toward irregularization in certain contexts, and are more in line with the first hypothesis. 
    
\end{abstract}

\noindent\textbf{Keywords:} phylogenetic comparative methods, 
morphological change, 
analogical change, 
paradigm leveling, 
Germanic

\section{Introduction}

Historical morphology focuses on the minimal meaning-bearing units in language and their change over time. By investigating how the mapping between meaning and form evolves in these fundamental units, we can better understand the processes that shape languages. Historical linguistics provides a framework for examining these transformations, which helps in studying the cognitive processes that underlie the linguistic system in the mind. The evolution of verbal paradigms and specifically the phenomenon of paradigm leveling, the process by which non-uniform patterns of lexical allomorphy are replaced with uniform ones, presents an interesting case study in this context. It allows us to explore questions about the balance between uniformity and non-uniformity in language, the role of analogy in linguistic change, and the factors that influence the direction and extent of morphological transformations. 
Examining specific instances of morphological change within well-documented language families provides insight into the broader mechanisms driving linguistic evolution and the organization of morphological systems. This raises questions about the nature and pathways of language change, challenging simplistic notions of unidirectional processes and inviting deeper exploration of the complex interplay between various linguistic and cognitive factors.

The evolution of strong verbs in the Germanic clade provides a concrete example of linguistic dynamics as discussed above and has long been a subject of interest in historical linguistics \citep{krygier1994disintegration, seebold1970vergleichendes, ringe2008proto, dammel_etal2010, desmet_vandevelde2019}. Strong verbs, 
characterized by irregular, phonologically unpredictable, lexeme-specific stem-vowel alternation patterns across different tense/aspect/mood (TAM) forms, 
serve as an intriguing case study for understanding 
the factors underlying the maintenance of irregular alternation patterns, as well as their loss through paradigm leveling. 
The study of paradigm leveling in these verbs offers specific insights into the nature and pathways of language change. 
While leveling as a linguistic process is well documented, the specific mechanisms driving this phenomenon remain a subject of debate \citep{Hill2007,garrett2008}. 
This raises important questions 
regarding the circumstances under which irregularity is introduced, maintained, and lost, and the extent to which system-internal factors as well as external social and cultural influences play a significant role. 
The evolution of strong verbs, occurring across multiple Germanic languages over centuries, provides a rich dataset with a relatively well-documented history for exploring these and related questions. 


This study explores how changes in the domain of present perfect constructions shape vowel alternation patterns in Germanic verbal stems. 
By examining this relationship, we demonstrate a broader linguistic principle: semantic changes systematically influence morphological preferences. 
The Germanic strong verb system offers a case study into how languages adjust their inflectional patterns in response to evolving grammatical meanings.

To study this relationship, we use phylogenetic comparative methods \citep{Garamszegi2014,Dunn2015,Cathcart2018modeling}, a family of methodologies modeling the evolutionary dynamics of biological, linguistic, and cultural traits. 
Our model is designed to investigate evolutionary dependencies between sets of linguistic variables or characters. 
Our approach allows us to evaluate differences in evolutionary dynamics under varying grammatical conditions. In our analysis, we examine the interaction between two linguistic features across the Germanic clade: the functional domain of present perfect constructions, and the vowel alternation patterns observed in 107 strong verbs across 14 modern and historical varieties. 
We employ a hierarchical phylogenetic model \citep{Cathcartetal2022Decoupling,jing2022phylogenetic,cathcart2024,cathcart2024exploring} intended to isolate large-scale, global trends in our data while accounting for variation at the level of individual data points, in a manner analogous to multilevel regression models \citep{gelman2007data}. 

Our model's findings indicate a relationship between the functional overlap of present perfect constructions with past tense constructions and the stability of certain vowel alternation patterns in verb stems. 
Our results shed light on the mechanisms that underlie this relationship: 
specifically, when perfect constructions overlap in the functional domain with past constructions, strong verbs are more likely to maintain existing patterns where the past tense and past participle 
share the same vowel, though they are not necessarily more likely to develop such patterns anew. This asymmetric relationship suggests that functional overlap between past participle and past forms acts as a conserving force rather than a driving force for morphological change. Our findings not only support the broader hypothesis that semantic and functional changes influence morphological patterns, but also provide a quantitative basis for understanding the directionality of such influences. This relationship contributes to our understanding of how languages alter or preserve characteristics of their morphological systems in response to evolving semantic and syntactic structures.

\section{Background}
\subsection{Germanic strong verbs}
The phenomenon of strong verb alternations can be traced back to the Proto-Indo-European (PIE) ablaut system, which involved systematic vowel changes to signal grammatical contrasts \citep[Ch. 2]{ringe2008proto}. As the Germanic branch evolved, these alternations became more regularized\footnote{For our purposes, ``regularity'' refers to absence or reduced presence of strong verbs' stem-vowel alternations.} in certain contexts, while in others, they underwent significant modification through analogical processes. Despite the differences between PIE and Proto-Germanic (PG) caused by the accumulation of changes over time, the heritage of PIE is observed in PG and its daughter languages. One example where this similarity can be observed is in verbal paradigms. 

To avoid the need to analyze the entire verbal paradigm, this study focuses on the principal parts of the strong verbs' paradigms as commonly used in traditional grammars. Principal parts are the specific forms of a verb that allow speakers to infer all other forms in the paradigm, given the set of affixal processes for each cell. In the case of Germanic languages, it is customary to exemplify the four principal parts of a strong verb by listing the following forms: the present infinitive, the third person singular past indicative, the third person plural past indicative, and the past participle \citep[264]{ringe2008proto}. Proto-Germanic strong verbs are those verbs that had different vowels across most of the principal parts.\footnote{According to \cite{ringe2008proto}, Proto-Germanic verbs of classes 1--3 only had the same vowel for the past indicative 3rd person plural and the past participle. Verbs of classes 5 and 6 had the same vowel in the infinitive and the past participle, while class 6 additionally has the same vowel for both past tense forms.} 

Condensing the entire verbal paradigm to a simpler representation of four forms helps alleviate the potential problem of data sparsity and is simpler from the point of view of the current analysis. At the same time, this simplification is justified by the morphomic nature of the concept of principal parts. Morphomes are abstract units of morphological structure that do not directly correspond to phonological or semantic features but reveal the organization and regularities within inflectional systems \citep{aronoff1994morphology}. For example, if a single stem is shared across a few TAM cells of a verb and the relationship between those cells is not immediately apparent, this regularity could be labeled as a morphome. An important property of morphomes is the notion of diachronic coherence \citep{maiden2018morphomes}. If a paradigm cell within a morphomic structure changes its form, all other cells that are part of the morphome tend to change as well. This diachronic coherence ensures that changes in the principal parts will systematically reflect across the entire verbal paradigm, making the analysis of principal parts a proxy for the analysis of the whole paradigm change. 

Throughout the evolution of PG, many strong verbs have been fully leveled, thus becoming ``weak'' verbs --- the term used to denote verbs that have no vowel alternations in their principal parts. Normally, weak verbs mark their TAM category by means of affixation, 
e.g., a dental ({\it -t}, {\it -d}) suffix in the past tense. An instance of such change can be found in the evolution of Old English to Modern English, where the historically strong verb \textit{help} became fully leveled, as shown in (\ref{old_english_level_ex}).

Partial leveling, another significant phenomenon in Germanic languages, involves the harmonization of some forms within a paradigm while leaving others with distinct characteristics. This process typically aims to reduce vowel alternations within a verb's inflectional paradigm by standardizing certain forms while maintaining others. For instance, in the transition from Middle High German to New High German, the verb \textit{geben} `to give' illustrates partial leveling, as in (\ref{mhg_partlevel_ex}). The partial leveling process is evident in the loss of the vowel alternations in the past tense forms. In this particular example, the past tense acquired a uniform vowel across all person/number paradigm cells, instead of having two different vowels in singular and plural.

\begin{exe}
\ex \textbf{Full leveling:} Old English > English \citep[55]{fertig2013analogy} \label{old_english_level_ex}\\
\\
h\underline{\bf e}lpan - h\underline{\bf ea}lp - h\underline{\bf u}lpon - (ġe)h\underline{\bf o}lpen > h\underline{\bf e}lp - h\underline{\bf e}lped - h\underline{\bf e}lped - h\underline{\bf e}lped\\
Infinitive - 3SG Past Tense - 3PL Past Tense - Past Participle \\
\end{exe}

\begin{exe}
\ex \textbf{Partial leveling:} Middle High German > New High German \citep{benecke1863geben}  \label{mhg_partlevel_ex}\\
\\
g\underline{\bf ë}ben - g\underline{\bf a}p - g\underline{\bf â}ben - geg\underline{\bf ë}ben > g\underline{\bf e}ben - g\underline{\bf a}b - g\underline{\bf a}ben - geg\underline{\bf e}ben\\
Infinitive - 3SG Past Tense - 3PL Past Tense - Past Participle \\
\end{exe}

The current study is, at its core, concerned with the different trajectories of leveling. As pointed out by \cite{garrett2008}, many theoretical models assume a preference for non-alternating paradigms. 
In historical linguistics, sound change is usually charged with the role of creating alternations in the first place which are at first morphophonemic (i.e., phonologically predictable) and may become morphologized (i.e., unpredicatable) following the operation of subsequent sound changes, whereas paradigm leveling (or analogical changes, more broadly speaking) eliminates the consequences of the sound change \citeyearpar[193]{Hock2021}. 
The fact that changes involving leveling are well documented (including those affecting stem alternations in the Germanic verb, which are due to Proto-Indo-European ablaut as well as Verner's law, a process of prosodically conditioned voicing) lead many to posit it as a major force in language change \citep{Anttila1977,Wurzel1984}. 
However, despite these ideas, \cite{MannStephen2022Cats} highlight the fact that inflectional classes are highly stable across millennia, which appears to be inconsistent with a simple preference for uniformity. The definition of inflectional classes is a multilayered one but, importantly, shared idiosyncrasies of the phonological changes in the process of inflecting a verb, such as stem-vowel alternation, are one of the characteristics of an inflectional class \citep{corbett2006prolegomena}. The introduction of inflectional classes adds a layer of complexity to the idea of uniformity, as there might be a preference not only for intraparadigmatic uniformity but also for intraclass uniformity, with the latter potentially being a stronger pressure and thus explaining the non-uniform paradigms found in the modern varieties of Germanic and elsewhere.

\subsection{Previous findings}


Recent scholarship has investigated the mechanisms responsible for analogical leveling in Germanic strong verbs. 
Two notable studies, by \cite{dammel_etal2010} and \cite{desmet_vandevelde2019}, investigate the trajectories of full and partial leveling in strong verbs across different varieties of the Germanic clade; both find a relationship between leveling patterns and grammatical meaning expressed by present perfect constructions.

\begin{figure}[!ht]
    \centering
    \begin{subfigure}[b]{0.4\textwidth}
        \centering
        \includegraphics[width=\textwidth]{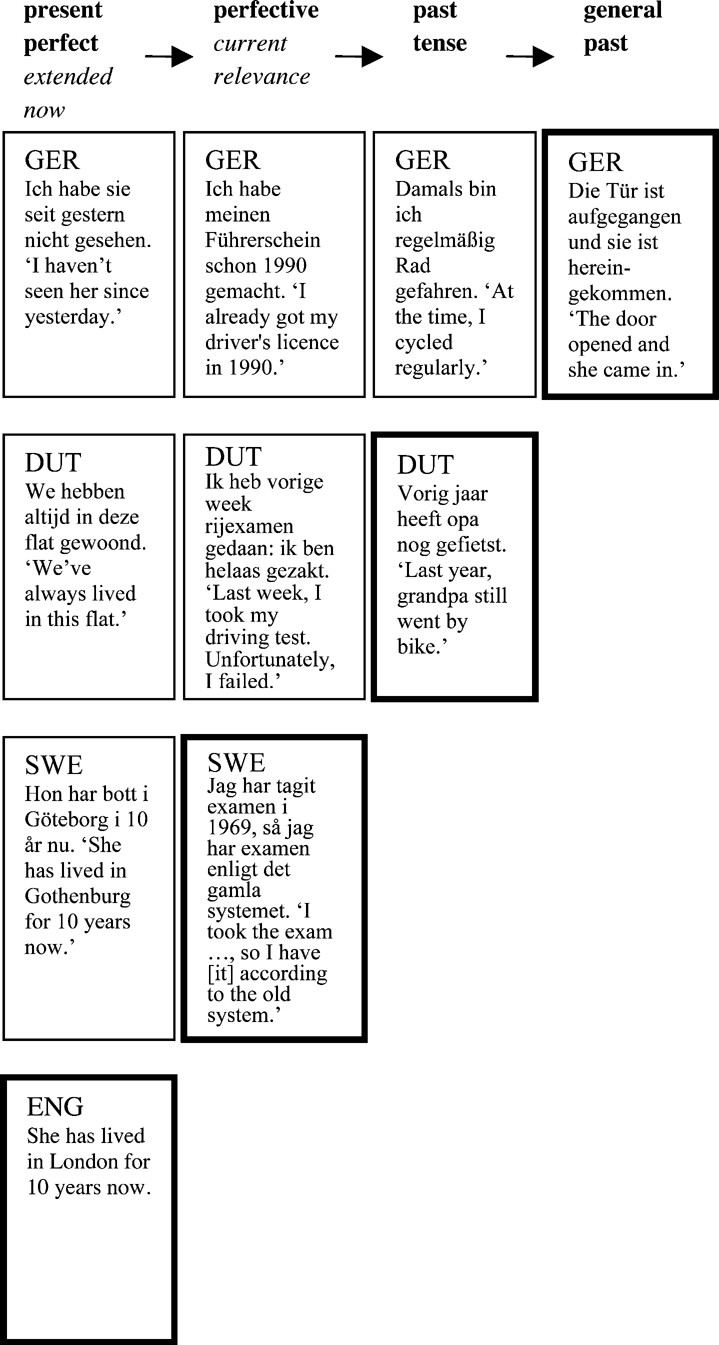}
        \caption{Functional expansion of the perfect \citep{dammel_etal2010}.}
        \label{fig:figure1}
    \end{subfigure}
    \hfill
    \begin{subfigure}[b]{0.5\textwidth}
        \centering
        \raisebox{3.25cm}{ 
        \includegraphics[width=\textwidth]{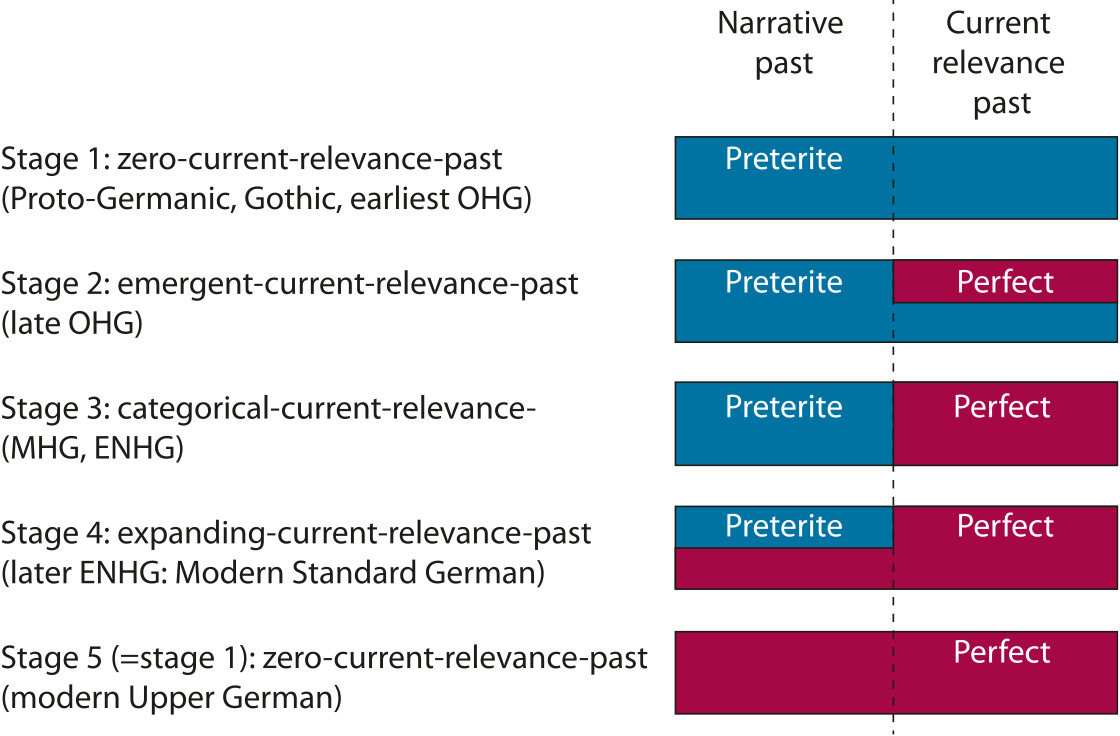}
        }
        \caption{Five stages in the development of the preterite and perfect \citep{seiler2022reichenbach}.}
        \label{fig:figure2}
    \end{subfigure}
    \caption{(a) \cite{dammel_etal2010} propose an explanation according to which the greater functional expansion of perfect leads to leveling in the stem vowel of the past participle; (b) \cite{desmet_vandevelde2019}, on the other hand, conclude that situations such as the one in Stage 2 and 4, where past tense and perfect overlap in their function, lead to a greater preference for partial leveling, with the past tense and past participle having the same stem vowel and the infinitive having a different one.}
    \label{fig:perfect}
\end{figure}


The present perfect expresses the continuing relevance of a past situation to the current point in time \citep[52]{comrie76}. As can be seen in Figure \ref{fig:figure2}, the present perfect construction does not require a separate form in order to express relevance to the present; this function may be taken by a simple past tense, where the semantics of current relevance is added by discourse. This is thought to be the case for Proto-Germanic and Gothic, and even some other historical and modern Germanic languages, but what sets the former two apart from these other languages is the fact that they do not have an alternative form that would uniquely encode perfect meaning. The analytic present perfect construction was an innovation of Northwest Germanic, which sets it apart from East Germanic, represented here by Gothic. The perfect originated via the grammaticalization of resultative constructions \citep{gillmann2016perfekt}. Figure \ref{fig:perfect} shows how in some Germanic languages, the meaning of the perfect was generalized so that the condition of being relevant to the present was dropped. While both \cite{dammel_etal2010} and \cite{desmet_vandevelde2019} concur on the existence of a relationship between present perfect constructions and leveling patterns, their proposed mechanisms differ in specific details. 

\cite{dammel_etal2010} conduct a survey of four Germanic languages: Modern High German, English, Swedish, and Dutch. Their findings align with their proposed mechanism, highlighting notable differences in the behavior of Swedish compared to the other languages. The authors observe that Swedish maintains a more prominent aspectual distinction between the past and the perfect, resulting in a higher relative frequency of past tense forms. This frequency difference, they argue, accounts for the unique pattern of ablaut leveling in Swedish, where a distinct past tense vowel is always preserved, and strong past tense forms show greater resistance to class change than strong past participles \citep[353]{dammel_etal2010}. The authors further elaborate that the influence of frequency is indirect, operating through its strong association with cognitive entrenchment \citep{divjak2015frequency}.


\cite{desmet_vandevelde2019} examine the weakening patterns of strong verbs in Dutch across various diachronic stages (Old Dutch, Middle Dutch, Early Modern Dutch, Modern Dutch) and to identify variables influencing the timing of a strong verb's transition to weak.\footnote{\citet[157]{desmet_vandevelde2019} clarify: ``We want to predict \textit{when} a verb became weak (if at all), not \textit{if} a verb became weak, given the time period the observation is drawn from."} Among several independent variables, vowel alternation patterns are included as a variable in their analysis. Their findings reveal that strong verbs with identical vowels in past tense and past participle forms tend to resist weakening significantly longer than those with distinct vowels across all principal parts. Conversely, strong verbs sharing the same vowel in present tense and past participle forms, but differing in the past tense form, exhibit the most rapid weakening according to their model. The authors emphasize that in Dutch, past tense and analytic perfect express the same temporal domain, suggesting that it is advantageous for a strong verb to differentiate present and past by employing distinct vowels for these temporal domains while maintaining consistency within each domain.\footnote{The precise nature of this advantage is not explicitly stated. As this explanation resembles the one form/one meaning principle \citep{oneformonemeaning,dressler2005word,vennemann1972phonetic}, we might expect similar rationales to apply here.} 
As such, this phenomenon can be linked to a more general observation regarding the persistence of motivated allomorphy that marks functionally important semantic oppositions \citep{Kurylowicz1945,manczak1957tendances,manczak1978lois,manczak1980laws,hooper1979}. 

Both of these studies, while providing valuable insights, share certain methodological limitations that potentially impact the 
broader applicability of their conclusions.
The scope of both studies is relatively narrow, focusing on a limited number of Germanic languages. \cite{dammel_etal2010} surveyed only four languages, while \cite{desmet_vandevelde2019} based their analysis only on Dutch and its ancestors. This limited scope raises questions about the generalizability of their findings to the rest of the Germanic language family and to language more broadly.
Because of this, the identified patterns and proposed mechanisms, while compelling within the context of the studied languages, may not be universally applicable. It remains unclear whether these patterns of vowel alternation and their relationship to present perfect constructions would hold true in other Germanic languages or in languages from different families with similar morphological features.

Furthermore, the methodology employed by \cite{dammel_etal2010} lacks rigorous inferential statistics to validate their hypothesis. Their conclusions primarily rest on descriptive observations from a limited sample of four languages. To illustrate their point, the authors present Figure \ref{fig:figure1}, depicting the functional breadth of perfect constructions, alongside distributions of vowel alternation types for each language. They note a correlation between increased functional breadth and a higher proportion of non-leveled vowel patterns in past participles. However, the authors acknowledge that English diverges from this proposed pattern. Despite its perfect constructions being unable to express narrative past, the distribution of vowel alternating patterns among strong verbs in English aligns more closely with German and Dutch than with Swedish, as their hypothesis would predict. While this counterexample does not necessarily invalidate their proposal, it does suggest a probabilistic relationship rather than a categorical one. This nuance underscores the need for a more robust inferential model to assess whether this expectation holds true, even as a general tendency. Finally, it is worth mentioning that \citeauthor{dammel_etal2010}'s (\citeyear{dammel_etal2010}) claim that functional expansion of the analytic perfect increases its frequency is an empirical question that remains open. While the authors show that past participles are more frequent than past tense forms in Dutch, its difference from Swedish is not striking.\footnote{The authors only collected verbs in past tense and in past participle forms. Their sum represents 100\%.} Dutch uses past participles in 54\% of cases in their corpora, while Swedish does so in 43\% of the forms. English uses the past participle forms only in 32\% of cases. This further emphasizes English's strange behavior under the explanation proposed by \cite{dammel_etal2010}.

The study by \cite{desmet_vandevelde2019} 
relies on historical data of a single language, Dutch, to draw inferences about the broader West Germanic group. This narrow focus raises questions about the applicability of their findings to other West Germanic languages. It remains unclear whether the observed relationship between analytic perfect constructions and vowel patterns is a Dutch-specific innovation or if it simply emerges in Dutch due to the presence of necessary preconditions that may or may not exist in other West Germanic languages. Furthermore, in addressing the broader question of directionality in paradigm leveling, the authors adopt a somewhat simplistic approach by focusing solely on full leveling, i.e., complete weakening of strong verbs. This narrow focus potentially obscures the nuanced pathways of weakening and fails to capture the intermediate stages of partial leveling that may occur. Such partial leveling processes could provide valuable insights into the gradual nature of morphological change and the factors influencing its progression.

These two studies differ slightly in their descriptions of how the functional broadness of the analytic perfect affects the preferred vowel alternation patterns. In this study, we will not be able to adjudicate the more minor points of difference between these analyses; instead, we motivate a method that can robustly establish whether or not this particular link between function and form exists, while accounting for multiple factors, such as phylogeny, idiosyncrasies of individual verbs, and different pathways of weakening. 
We are additionally able to shed light on the dynamics of diachronic change underlying this association. 


\section{Data}
\label{data}

Our study relies on a comprehensive dataset of Germanic verbal forms, collected and processed to ensure broad coverage of verbs with vowel alternation patterns across 14 languages of the Germanic language family: Icelandic, English, Faroese, Frisian, Old Saxon, Danish, Low German, Old English, German, Norwegian Bokmål, Dutch, Gothic, Old High German, and Swedish. This set, although incomplete, ensures good coverage of all the sub-families of the Germanic clade.

Our data collection process involved several steps. First, 
we automatically extracted relevant Germanic verbal forms from Wiktionary (\url{wiktionary.org}), a collaborative online dictionary.\footnote{Our data extraction was carried out on 16 September 2024.} This source provided a wide range of verbal forms across various Germanic languages, including both modern and historical varieties. 
After that, to complement and cross-validate the Wiktionary data, we incorporated verbal forms found in UniMorph \citep{unimorph} which contains several Germanic languages. This addition helped ensure a more comprehensive coverage and allowed for verification of the scraped Wiktionary data, in addition to filling in gaps for some of the verbs. To avoid an overly extensive manual search of the data, we limited ourselves to verbs that were present in at least 80\% of languages after this automated procedure, leaving us with a final dataset of 107 strong verbs. Since not all 107 verbs were successfully scraped automatically,\footnote{In some cases the 
script 
extracted the infinitive form of the verb but not the inflected forms and in some cases, no form was extracted at all.} we had to manually search for those verbs and/or their inflected forms in Wiktionary. Additionally, for some Old High German forms, we used an online Old High German dictionary published by the Saxon Academy of Science (\citeyear{awb_online}). After concluding this process, the forms were validated against \citeauthor{kroonen2013etymological}'s etymological dictionary (2013). 

\begin{figure}[!ht]
    \centering    \includegraphics[width=.75\linewidth]{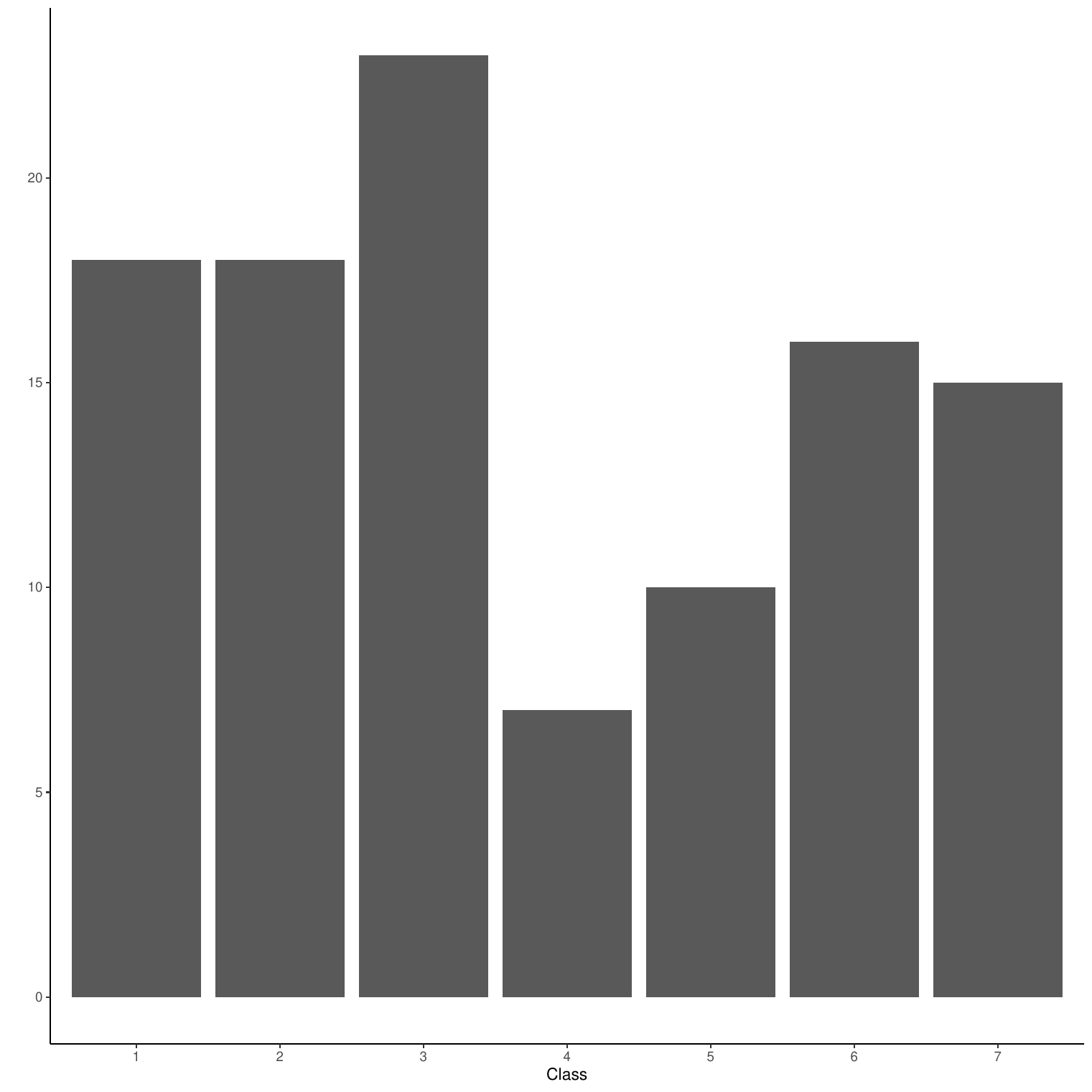}
    \caption{The distribution of classes (cf.~\citet[112-120]{mhg_dict}) for the 107 verbs in the dataset. For classes 5, 6 and 7, both strong verbs and \textit{*j}-present strong verbs are counted. For class 7, all subclasses (e.g., 7a) are included  in the count.}
    \label{fig:class_distribution}
\end{figure}

\begin{figure}[!ht]
    \centering    
    \begin{adjustbox}{max totalsize={\linewidth}{\linewidth},center}
\begin{tikzpicture}[x=1pt,y=1pt]
\definecolor{fillColor}{RGB}{255,255,255}
\path[use as bounding box,fill=fillColor,fill opacity=0.00] (0,0) rectangle (867.24,729.93);
\begin{scope}
\path[clip] (  0.00,  0.00) rectangle (867.24,729.93);
\definecolor{drawColor}{RGB}{255,255,255}
\definecolor{fillColor}{RGB}{255,255,255}

\path[draw=drawColor,line width= 0.6pt,line join=round,line cap=round,fill=fillColor] (  0.00,  0.00) rectangle (867.24,729.93);
\end{scope}
\begin{scope}
\path[clip] ( 31.71, 84.16) rectangle (861.74,724.43);
\definecolor{fillColor}{RGB}{255,255,255}

\path[fill=fillColor] ( 31.71, 84.16) rectangle (861.74,724.43);
\definecolor{fillColor}{gray}{0.35}

\path[fill=fillColor] ( 40.48,113.26) rectangle ( 93.09,695.32);

\path[fill=fillColor] ( 98.93,113.26) rectangle (151.54,567.27);

\path[fill=fillColor] (157.39,113.26) rectangle (209.99,509.06);

\path[fill=fillColor] (215.84,113.26) rectangle (268.45,509.06);

\path[fill=fillColor] (274.29,113.26) rectangle (326.90,357.73);

\path[fill=fillColor] (332.74,113.26) rectangle (385.35,276.24);

\path[fill=fillColor] (391.20,113.26) rectangle (443.80,241.31);

\path[fill=fillColor] (449.65,113.26) rectangle (502.26,183.11);

\path[fill=fillColor] (508.10,113.26) rectangle (560.71,171.47);

\path[fill=fillColor] (566.55,113.26) rectangle (619.16,159.83);

\path[fill=fillColor] (625.01,113.26) rectangle (677.61,159.83);

\path[fill=fillColor] (683.46,113.26) rectangle (736.07,136.54);

\path[fill=fillColor] (741.91,113.26) rectangle (794.52,124.90);

\path[fill=fillColor] (800.36,113.26) rectangle (852.97,113.26);
\end{scope}
\begin{scope}
\path[clip] (  0.00,  0.00) rectangle (867.24,729.93);
\definecolor{drawColor}{RGB}{0,0,0}

\path[draw=drawColor,line width= 0.6pt,line join=round] ( 31.71, 84.16) --
	( 31.71,724.43);
\end{scope}
\begin{scope}
\path[clip] (  0.00,  0.00) rectangle (867.24,729.93);
\definecolor{drawColor}{gray}{0.30}

\node[text=drawColor,anchor=base east,inner sep=0pt, outer sep=0pt, scale=  0.88] at ( 26.76,110.23) {0};

\node[text=drawColor,anchor=base east,inner sep=0pt, outer sep=0pt, scale=  0.88] at ( 26.76,226.64) {10};

\node[text=drawColor,anchor=base east,inner sep=0pt, outer sep=0pt, scale=  0.88] at ( 26.76,343.06) {20};

\node[text=drawColor,anchor=base east,inner sep=0pt, outer sep=0pt, scale=  0.88] at ( 26.76,459.47) {30};

\node[text=drawColor,anchor=base east,inner sep=0pt, outer sep=0pt, scale=  0.88] at ( 26.76,575.88) {40};

\node[text=drawColor,anchor=base east,inner sep=0pt, outer sep=0pt, scale=  0.88] at ( 26.76,692.29) {50};
\end{scope}
\begin{scope}
\path[clip] (  0.00,  0.00) rectangle (867.24,729.93);
\definecolor{drawColor}{gray}{0.20}

\path[draw=drawColor,line width= 0.6pt,line join=round] ( 28.96,113.26) --
	( 31.71,113.26);

\path[draw=drawColor,line width= 0.6pt,line join=round] ( 28.96,229.67) --
	( 31.71,229.67);

\path[draw=drawColor,line width= 0.6pt,line join=round] ( 28.96,346.09) --
	( 31.71,346.09);

\path[draw=drawColor,line width= 0.6pt,line join=round] ( 28.96,462.50) --
	( 31.71,462.50);

\path[draw=drawColor,line width= 0.6pt,line join=round] ( 28.96,578.91) --
	( 31.71,578.91);

\path[draw=drawColor,line width= 0.6pt,line join=round] ( 28.96,695.32) --
	( 31.71,695.32);
\end{scope}
\begin{scope}
\path[clip] (  0.00,  0.00) rectangle (867.24,729.93);
\definecolor{drawColor}{RGB}{0,0,0}

\path[draw=drawColor,line width= 0.6pt,line join=round] ( 31.71, 84.16) --
	(861.74, 84.16);
\end{scope}
\begin{scope}
\path[clip] (  0.00,  0.00) rectangle (867.24,729.93);
\definecolor{drawColor}{gray}{0.20}

\path[draw=drawColor,line width= 0.6pt,line join=round] ( 66.78, 81.41) --
	( 66.78, 84.16);

\path[draw=drawColor,line width= 0.6pt,line join=round] (125.24, 81.41) --
	(125.24, 84.16);

\path[draw=drawColor,line width= 0.6pt,line join=round] (183.69, 81.41) --
	(183.69, 84.16);

\path[draw=drawColor,line width= 0.6pt,line join=round] (242.14, 81.41) --
	(242.14, 84.16);

\path[draw=drawColor,line width= 0.6pt,line join=round] (300.59, 81.41) --
	(300.59, 84.16);

\path[draw=drawColor,line width= 0.6pt,line join=round] (359.05, 81.41) --
	(359.05, 84.16);

\path[draw=drawColor,line width= 0.6pt,line join=round] (417.50, 81.41) --
	(417.50, 84.16);

\path[draw=drawColor,line width= 0.6pt,line join=round] (475.95, 81.41) --
	(475.95, 84.16);

\path[draw=drawColor,line width= 0.6pt,line join=round] (534.41, 81.41) --
	(534.41, 84.16);

\path[draw=drawColor,line width= 0.6pt,line join=round] (592.86, 81.41) --
	(592.86, 84.16);

\path[draw=drawColor,line width= 0.6pt,line join=round] (651.31, 81.41) --
	(651.31, 84.16);

\path[draw=drawColor,line width= 0.6pt,line join=round] (709.76, 81.41) --
	(709.76, 84.16);

\path[draw=drawColor,line width= 0.6pt,line join=round] (768.22, 81.41) --
	(768.22, 84.16);

\path[draw=drawColor,line width= 0.6pt,line join=round] (826.67, 81.41) --
	(826.67, 84.16);
\end{scope}
\begin{scope}
\path[clip] (  0.00,  0.00) rectangle (867.24,729.93);
\definecolor{drawColor}{gray}{0.30}

\node[text=drawColor,rotate= 45.00,anchor=base east,inner sep=0pt, outer sep=0pt, scale=  0.88] at ( 71.07, 74.92) {West Frisian};

\node[text=drawColor,rotate= 45.00,anchor=base east,inner sep=0pt, outer sep=0pt, scale=  0.88] at (129.52, 74.92) {Old High German};

\node[text=drawColor,rotate= 45.00,anchor=base east,inner sep=0pt, outer sep=0pt, scale=  0.88] at (187.97, 74.92) {Gothic};

\node[text=drawColor,rotate= 45.00,anchor=base east,inner sep=0pt, outer sep=0pt, scale=  0.88] at (246.43, 74.92) {Low German};

\node[text=drawColor,rotate= 45.00,anchor=base east,inner sep=0pt, outer sep=0pt, scale=  0.88] at (304.88, 74.92) {Faroese};

\node[text=drawColor,rotate= 45.00,anchor=base east,inner sep=0pt, outer sep=0pt, scale=  0.88] at (363.33, 74.92) {Danish};

\node[text=drawColor,rotate= 45.00,anchor=base east,inner sep=0pt, outer sep=0pt, scale=  0.88] at (421.79, 74.92) {Old Saxon};

\node[text=drawColor,rotate= 45.00,anchor=base east,inner sep=0pt, outer sep=0pt, scale=  0.88] at (480.24, 74.92) {Icelandic};

\node[text=drawColor,rotate= 45.00,anchor=base east,inner sep=0pt, outer sep=0pt, scale=  0.88] at (538.69, 74.92) {Norwegian};

\node[text=drawColor,rotate= 45.00,anchor=base east,inner sep=0pt, outer sep=0pt, scale=  0.88] at (597.14, 74.92) {German};

\node[text=drawColor,rotate= 45.00,anchor=base east,inner sep=0pt, outer sep=0pt, scale=  0.88] at (655.60, 74.92) {Swedish};

\node[text=drawColor,rotate= 45.00,anchor=base east,inner sep=0pt, outer sep=0pt, scale=  0.88] at (714.05, 74.92) {Old English};

\node[text=drawColor,rotate= 45.00,anchor=base east,inner sep=0pt, outer sep=0pt, scale=  0.88] at (772.50, 74.92) {Dutch};

\node[text=drawColor,rotate= 45.00,anchor=base east,inner sep=0pt, outer sep=0pt, scale=  0.88] at (830.95, 74.92) {English};
\end{scope}
\begin{scope}
\path[clip] (  0.00,  0.00) rectangle (867.24,729.93);
\definecolor{drawColor}{RGB}{0,0,0}

\node[text=drawColor,rotate= 90.00,anchor=base,inner sep=0pt, outer sep=0pt, scale=  1.10] at ( 13.08,404.29) {Missing verbs};
\end{scope}
\end{tikzpicture}
    \end{adjustbox}
    \caption{The distribution of missing verbs per language, out of a total of 107.}
    \label{fig:missing_per_lang}
\end{figure}
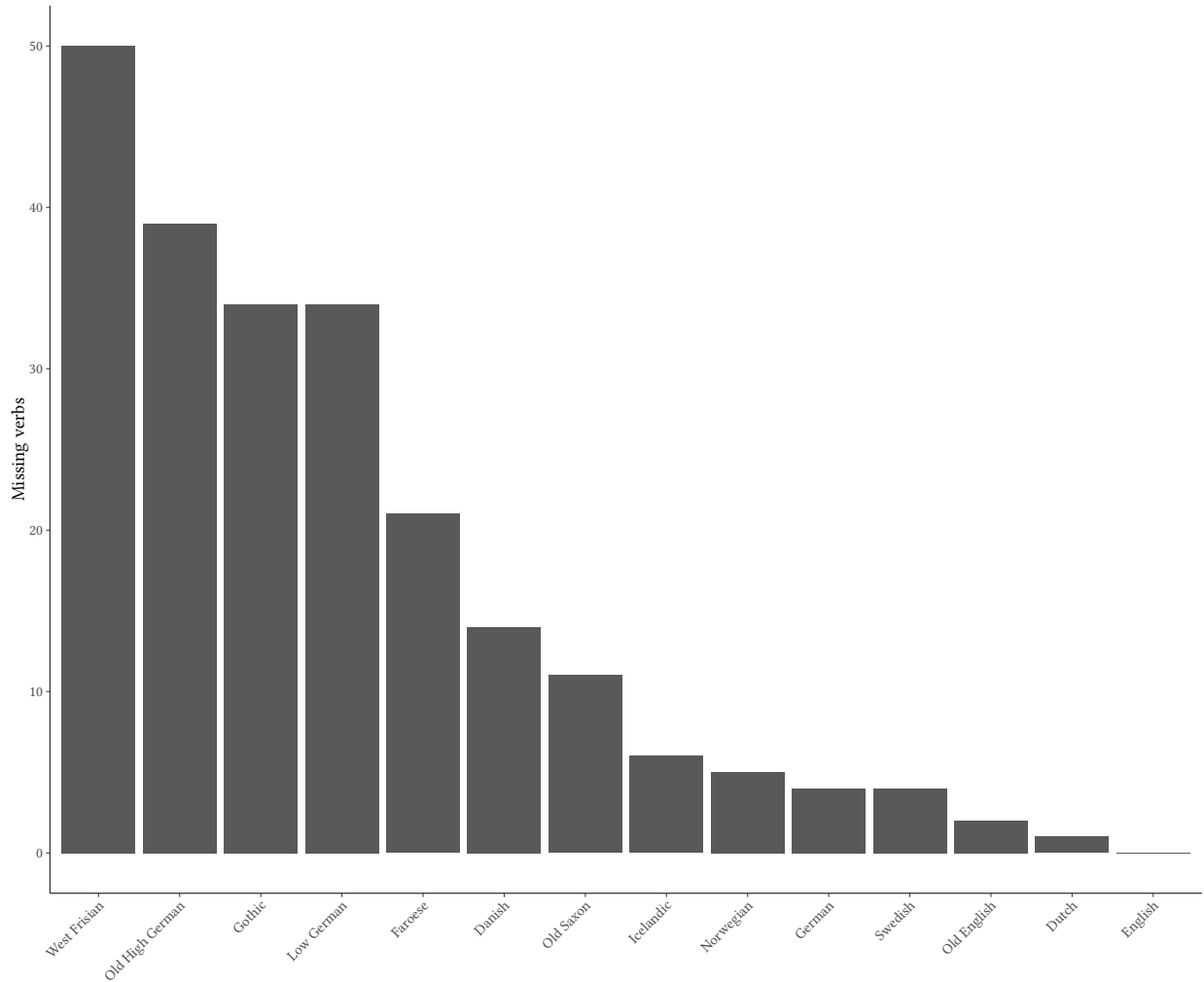

As illustrated in Figure \ref{fig:class_distribution}, our dataset includes all 7 verb classes found in Proto-Germanic. Class 4 appears less frequently in our data, reflecting its lower occurrence in Wiktionary. Overall, the distribution of verb classes in our dataset aligns roughly with their proportions in Wiktionary. Analyses of missing data points per language, shown in Figure \ref{fig:missing_per_lang}, display a clear pattern. Minority languages, such as Low German and West Frisian, have a high number of missing forms. The same is true of old languages, like Old High German and Gothic. However, no language is missing more than 50\% of the verbs and in 10 out of 14 languages less than 20\% of the forms are missing. 
We mark a verb as dead rather than missing in a given language if it is marked as obsolete or given a similar label in Wiktionary. 



Following data collection, we coded each verb according to its vowel alternation pattern. According to traditional scholarship \cite[72--74]{mhg_dict}, Germanic languages have four principal parts: the infinitive, 
the third person singular past indicative, the first person plural past indicative, and the past participle (``supine'' in the North Germanic tradition). 
However, for this study, we simplified the coding to three principal parts, for two reasons: reduction of the state space, and modern language patterns. Using four principal parts would result in a large and computationally intractable state space for the phylogenetic model. In phylogenetic comparative models, an excessively large state space can lead to computational intractability, overfitting, and difficulties in parameter estimation. 
Considering that we have 107 characters, it is likely that our models would not achieve convergence across chains (a crucial property of Bayesian models) if we had opted for four principal parts, limiting their explanatory power. 
Apart from computational considerations, many modern Germanic languages have leveled their past tense forms, using the same vowel for both third person singular and first person plural. As previously mentioned, \cite{dammel_etal2010} and \cite{desmet_vandevelde2019} also used only three principal parts, so this makes our results directly comparable to theirs. 

With respect to our coding scheme, we followed the coding decisions used in previous literature, as these represent an optimal trade-off between the complexity found in the paradigm's organization and the feasibility of the analysis. In our coding procedure, we only analyze the roots of the verb, ignoring inflectional affixes. If the vowel found in the root of the form differs from the one found in other paradigm cell(s), it receives a different letter coding. The order of letters corresponds to infinitive, past tense form and past participle, respectively. 
Explicit illustrations of coded patterns can be found below: 
\begin{itemize}

    \item \textbf{ABC}: Three different vowels in each of the three principal parts' forms. This pattern was the most frequent one in Proto-Germanic. Within our coding scheme it could be considered the most complex one, as it has both high enumerative (i.e., number of allomorphs) and integrative (i.e., predictive relationships between allomorphs) complexity \citep[cf.][]{AckermanMalouf2013}.
    
    Modern High German: tr\underline{\bf i}nken – tr\underline{\bf a}nk – getr\underline{\bf u}nken
    
    \item \textbf{ABB}: Same vowel in past and past participle. According to \citeauthor{dammel_etal2010}'s (\citeyear{dammel_etal2010}) calculations, this pattern is the most frequent one in Dutch, Modern High German and English. \cite{desmet_vandevelde2019} show an effect where Old Dutch's verbs with the ABB pattern were not fully leveled for a longer time than other patterns.
    
    Dutch: br\underline{\bf e}ngen – br\underline{\bf a}cht – gebr\underline{\bf a}cht

    \item \textbf{AAA}: Same vowel throughout the paradigm. Verbs with the AAA pattern are called  `weak' in traditional philological terms. We expect this pattern to rarely change into any of the other ones. Fully leveled verbs are frequently inflected using a productive rule, for example, by using a dental suffix for the past tense forms. Despite that, there are individual examples of transitions from AAA to ABB, such as past tense and past participle of `dive' becoming `dove', instead of the previous `dived'. 
    
    English: h\underline{\bf e}lp – h\underline{\bf e}lped – h\underline{\bf e}lped

    \item \textbf{ABA}: Same vowel in infinitive and past participle, but a different one in the past tense form. Same as in the case of \citeauthor{dammel_etal2010}'s (\citeyear{dammel_etal2010}) data, this pattern is most frequently found in the North Germanic clade. In addition to that, we find the ABA pattern frequently in older languages, such as Gothic and Old English. This pattern is also typical for certain classes of Proto-Germanic verbs, although it is less frequent than ABC.

    Swedish: skr\underline{\bf i}da - skr\underline{\bf e}d - skr\underline{\bf i}dit

    \item \textbf{AAB}: Same vowel in the infinitive and past tense, and a different one in the past participle. This pattern is extremely infrequent. It is difficult to find an explanation as to why it arose in cases where it did, but given the number of verbs that have this pattern, we could suggest that it is an example of idiosyncratic development in a limited number of verbs. 

    Icelandic: s\underline{\bf ö}kkva - s\underline{\bf ö}kk - s\underline{\bf o}kkið
    
\end{itemize}

\begin{figure}[!ht]
    \centering    \includegraphics[width=\linewidth]{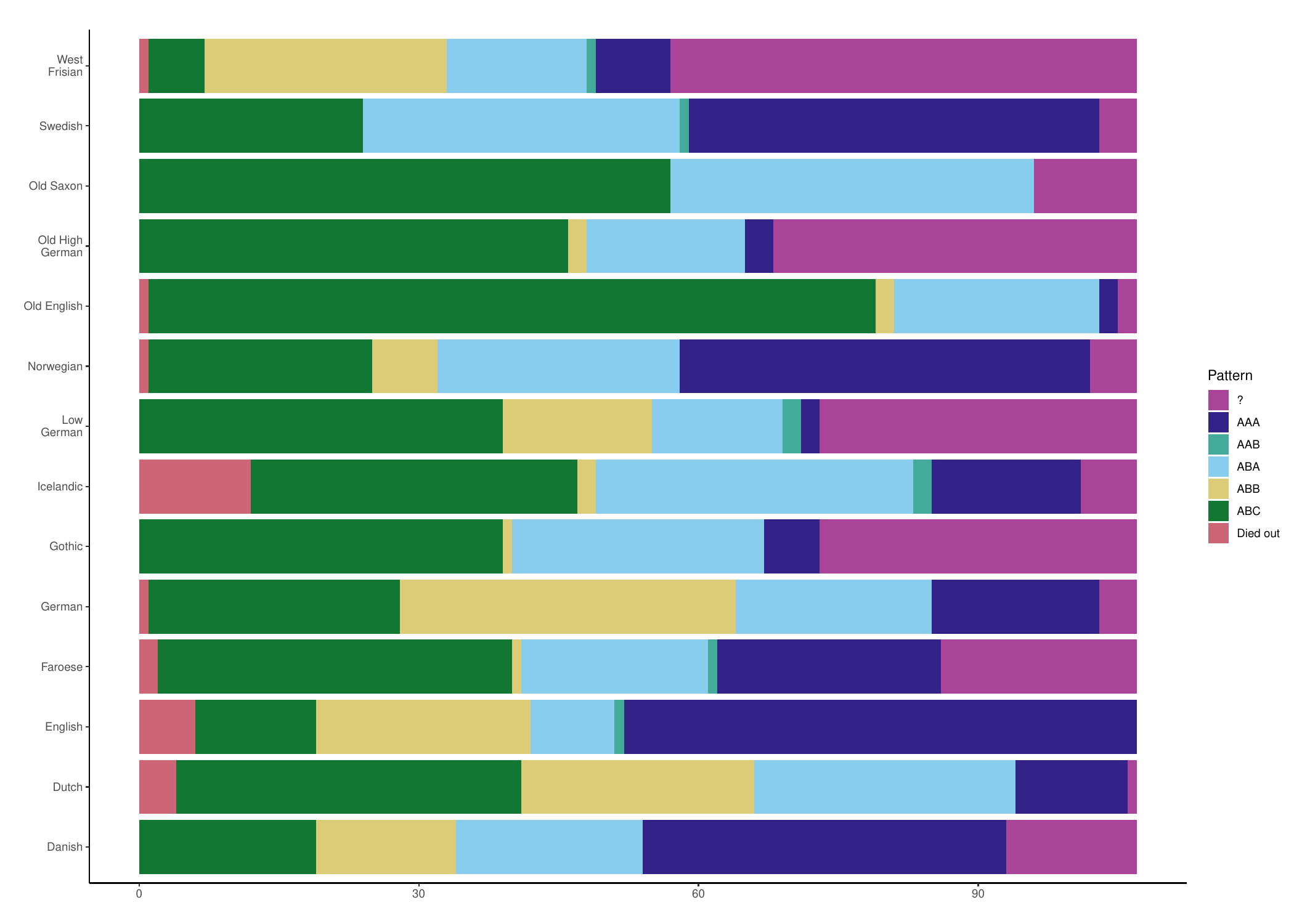}
    \caption{The distribution of coded vowel alternation patterns for each language.}
    \label{fig:pattern_distribution}
\end{figure}

The distribution of coded vowel alternation patterns for each language is displayed in Figure \ref{fig:pattern_distribution}. It is worth stressing that this coding scheme is intended to refer to stem similarities and differences in verbal paradigms on a purely abstract, verb-internal basis, and that there is not necessarily a meaningful relationship between alternants marked with the same letter across different verb paradigms. 
It is also worth noting that while in theory changes between alternation patterns could stem from regular sound changes such as mergers or conditioned splits, this is generally not the case for the Germanic data we analyze --- transitions between alternation patterns can be reliably taken to represent analogical morphological changes.

We analyze the dynamics of diachronic changes between alternation patterns at the verb level using a phylogenetic model. 
In our case, this requires the use of a phylogenetic timed tree of the Germanic languages, representing the relatedness and divergence between the languages in our sample. 
We use the Germanic clade within the Indo-European phylogenetic tree sample of \cite{chang2015} for our analyses.\footnote{We make the conscious decision not to base our analyses on a recent comprehensive phylogenetic study of the Indo-European languages \citep{heggarty2023}, which infers phylogenies from lexical data alone without clade constraints representing 
even uncontroversial 
subgroup-defining morphological and phonological innovations shared among closely related languages. While this method in many cases recapitulates properties of the Indo-European tree topology that are in line with received wisdom, we find that key aspects of the Germanic clade in the main published results are highly problematic. In particular, the authors report strong support for an unrealistically late date of divergence between modern Frisian, Dutch, and High German speech varieties, which form a subgroup to the exclusion of closely related archaic speech varieties (e.g., Old High German). For practical purposes, this effectively forces our stochastic character mapping procedure (see below) to reconstruct a single, recent, synapomorphic semantic shift of the present perfect construction to express narrative past, which is contradicted by the textual record \citep{seiler2022reichenbach}, and would potentially give rise to misleading results at the relatively narrow phylogenetic scope that we take for this paper.}

This phylogeny lacks certain taxa for which we were able to collect vowel alternation patterns, namely Low German and Old Saxon. To avoid losing data, these taxa were added to the trees manually, while preserving uncertainty about the topology and branch lengths. 
We first added the Old Saxon taxon. It was added as a sister node to the MRCA of Old High German and Modern High German. This created a topology where Old Saxon and its descendant, Low German, are closely related to the High German clade but do not have to undo the effects of the High German Consonant Shift. In terms of branch length, Old Saxon's age was again randomly generated, but it was centered around the same value as Old High German. Old Saxon manuscripts tend to be slightly older than Old High German manuscripts but their age difference is not major; hence, we assume that their age should be similar. After adding a branch for Old Saxon, 
we added a sibling branch for Low German to this branch, putting these two taxa into a close relationship that is effectively one of near-direct descent. 
The Low German branch length has been adjusted accordingly to reflect that it is a modern language. 

\begin{figure}[!ht]
    \centering    
    \includegraphics[width=170mm, angle=90]{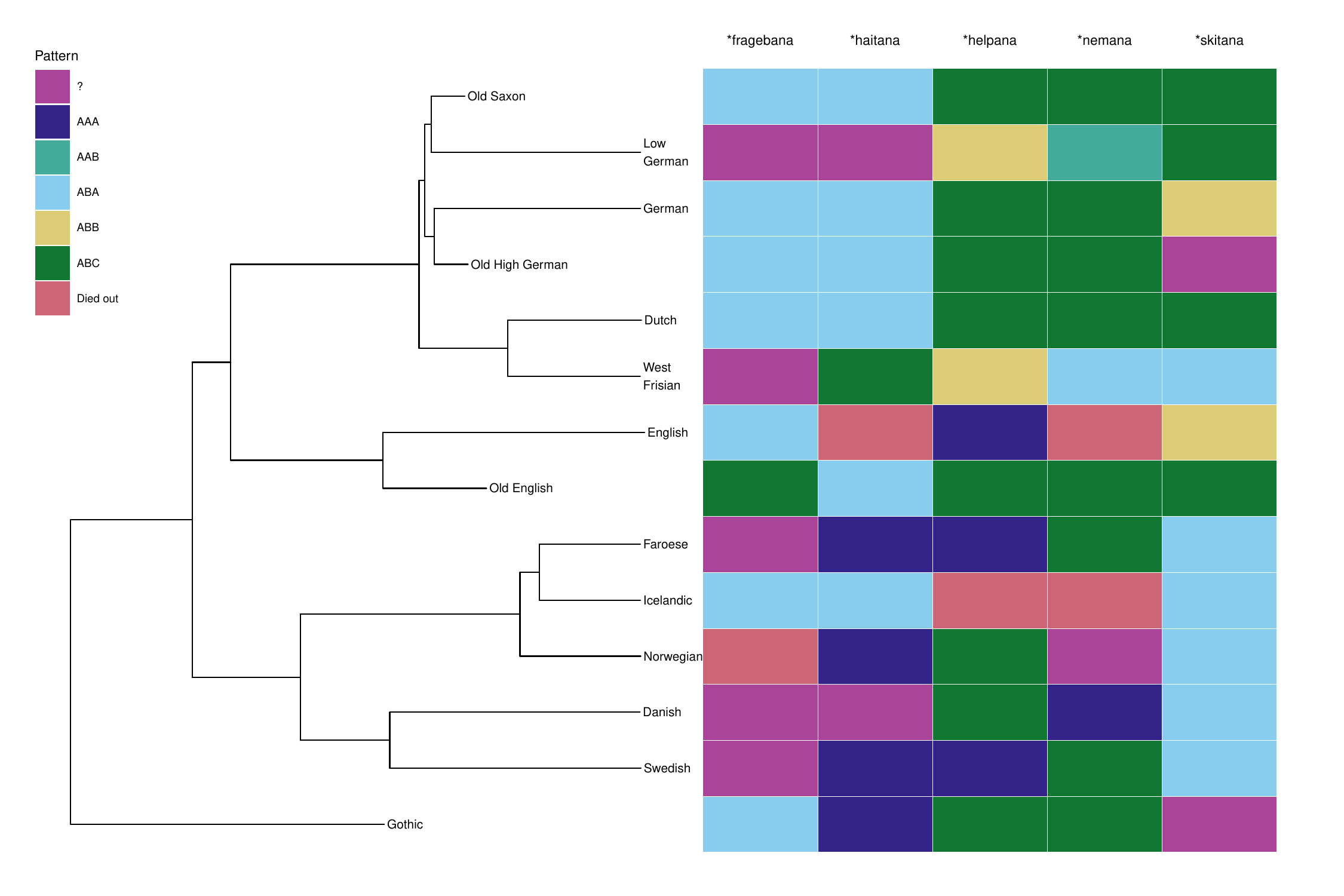}
    \caption{A sample of verbs and their coded patterns mapped to a Maximum Clade Credibility tree from \cite{chang2015}.}
    \label{fig:tree_char}
\end{figure}

A visualization of selected character data mapped onto a tree can be found in Figure \ref{fig:tree_char}. It should be stressed that this is a single tree from the tree samples we use; we account for phylogenetic uncertainty by running our model over multiple such trees.

Finally, the data for functional expansion of the analytic perfect were collected. We operationalized this expansion as a binary character. This character answers the question ``Was the present perfect construction expanded to be used for expressing narrative past?". Because of that, in upcoming sections, we will refer to the two states of this character as either {\sc extended} (E), or {\sc non-extended} (N). For the first case, a prime example is Modern High German, whereas English is a representative of a language where present perfect constructions were not extended. Both of these states are shown in Example \ref{examples:tam}.

\begin{exe}
    \ex The following sentences are calques in English and German, with the only difference being word order. \label{examples:tam}
        \begin{xlist}
            \ex \textbf{German}, extended ({\sc e}) perfect constructions:
            \\ Ich aß gestern Pasta $\equiv$ Ich habe gestern Pasta gegessen \label{ex:german}
            \\
            \ex \textbf{English}, non-extended ({\sc n}) perfect constructions: 
            \\ I ate pasta yesterday $\not\equiv$ ??I have eaten pasta yesterday \label{ex:eng}
        \end{xlist}
\end{exe}

The data for this character for Old High German and Low German came from \cite{seiler2022reichenbach}, for Old English and Old Saxon from \cite{macleod_2012}, while the rest of the languages were coded according to \cite{Weber2023}. 

\section{Methods}

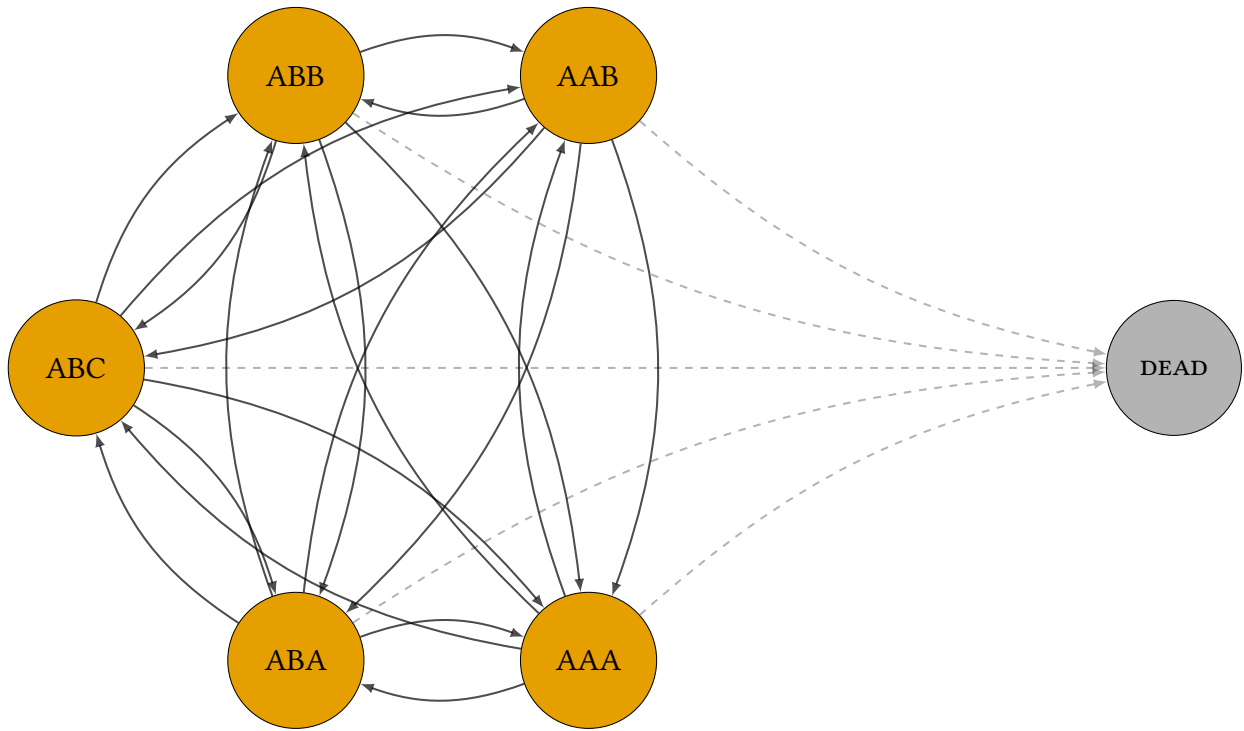
\begin{figure}[!ht]
    \centering
    \begin{adjustbox}{max totalsize={\linewidth}{\linewidth},center}
\definecolor{myorange}{HTML}{E69F00} 
\begin{tikzpicture}[node distance=2cm,->,>=latex,auto,
  every edge/.append style={thick}]
\def\s{2}
\centering
\draw (3*\s,1*\s) node[circle,minimum size=1cm,text width=1.5cm,align=center,draw,fill=myorange] (ABA) {ABA};
\draw (1.5*\s,3*\s) node[circle,minimum size=1cm,text width=1.5cm,align=center,draw,fill=myorange] (ABC) {ABC};
\draw (3*\s,5*\s) node[circle,minimum size=1cm,text width=1.5cm,align=center,draw,fill=myorange] (ABB) {ABB};
\draw (5*\s,1*\s) node[circle,minimum size=1cm,text width=1.5cm,align=center,draw,fill=myorange] (AAA) {AAA};
\draw (9*\s,3*\s) node[circle,minimum size=1cm,text width=1.5cm,align=center,draw,fill=gray!60] (DEAD) {{\sc dead}};
\draw (5*\s,5*\s) node[circle,minimum size=1cm,text width=1.5cm,align=center,draw,fill=myorange] (AAB) {AAB};

\path (AAA) edge[bend left=20, opacity=0.7] (AAB);
\path (AAA) edge[bend left=20, opacity=0.7] (ABB);
\path (AAA) edge[bend left=20, opacity=0.7] (ABA);
\path (AAA) edge[bend left=20, opacity=0.7] (ABC);
\path (AAA) edge[bend left=15, style=dashed, opacity=0.3] (DEAD);

\path (AAB) edge[bend left=20, opacity=0.7] (AAA);
\path (AAB) edge[bend left=20, opacity=0.7] (ABB);
\path (AAB) edge[bend left=20, opacity=0.7] (ABA);
\path (AAB) edge[bend left=20, opacity=0.7] (ABC);
\path (AAB) edge[bend right=15, style=dashed, opacity=0.3] (DEAD);

\path (ABB) edge[bend left=20, opacity=0.7] (AAB);
\path (ABB) edge[bend left=20, opacity=0.7] (ABA);
\path (ABB) edge[bend left=20, opacity=0.7] (AAA);
\path (ABB) edge[bend left=20, opacity=0.7] (ABC);
\path (ABB) edge[bend right=15, style=dashed, opacity=0.3] (DEAD);

\path (ABA) edge[bend left=20, opacity=0.7] (AAB);
\path (ABA) edge[bend left=20, opacity=0.7] (ABB);
\path (ABA) edge[bend left=20, opacity=0.7] (AAA);
\path (ABA) edge[bend left=20, opacity=0.7] (ABC);
\path (ABA) edge[bend left=15, style=dashed, opacity=0.3] (DEAD);

\path (ABC) edge[bend left=20, opacity=0.7] (AAB);
\path (ABC) edge[bend left=20, opacity=0.7] (ABB);
\path (ABC) edge[bend left=20, opacity=0.7] (ABA);
\path (ABC) edge[bend left=20, opacity=0.7] (AAA);
\path (ABC) edge[style=dashed, opacity=0.3] (DEAD);

\end{tikzpicture}
\end{adjustbox}
    \caption{Schema of a CTM model of evolution of stem alternation patterns in a verb form. During the course of its evolution over phylogenetic lineages of the Germanic subgroup, a verb form can transition between five stem alternation patterns (states represented by orange graph nodes, transitions represented by solid arrows) or can die out (transitions represented by dashed arrows). The state {\sc dead} is absorbing; while transitions to {\sc dead} can occur on divergent phylogenetic branches, such transitions are irreversible.}
    \label{fig:CTM}
\end{figure}


Under the phylogenetic comparative method that we use to analyze the evolution of stem alternations in Germanic, stem alternation patterns evolve over a phylogeny according to a continuous-time Markov (CTM) process (cf.\ \citealt{Dunn2011,haynie2016phylogenetic,Dunn2017,Cathcart2018modeling,Cathcart2020,shirtz2021evolutionary}, as well as \citealt{van2021markov}, where time-series change in verbal morphology is modeled with a CTM chain). 
Under this stochastic process, a system undergoes transitions between different states, as schematized in Figure \ref{fig:CTM}. 
Between-state transitions occur according to frequencies whose expected values are represented by transition rates; as there are five different states, a schema like that shown would have $5 ~\text{states} \times 4 ~\text{transitions (to other states)} ~+ 1 ~\text{death} ~\text{rates} = 21~\text{parameters}$. 
These parameters represent transitions between states as well as transitions to the state {\sc dead}, the last of which is state independent. 
A CTM chain has a stationary or equilibrium distribution, which represents the proportion of time that the system is expected to be in different states as time approaches infinity; this distribution can be interpreted as the long-term preference for particular states. 
In our study, we allow the basic CTM process shown in Figure \ref{fig:CTM} --- specifically, rates for transitions represented by solid lines --- to vary across regimes represented by different TAM states, yielding a basic model with $41$ parameters ($5 ~\text{states} \times 4 ~\text{transitions} \times 2~\text{regimes} ~+ 1 ~\text{death} ~\text{rate}$). 
This allows us to investigate whether there are different propensities for certain stem alternation patterns depending on whether present perfect expresses narrative past. 

In the model setting we employ, the evolution of the analytic perfect constructions in Germanic is treated as a given, observed entity on which the evolution of verb-level patterns is conditioned, rather than co-inferring properties of the evolution of all variables of interest. 
In our model, which lineages (in this case branches or segments of branches) belong to which regime is given information that serves as input to our model. 
So-called ``painting'' of regimes onto trees \citep{beaulieu2012modeling} is standard practice for complex models like the one we use. 

\begin{figure}
    \centering
    \includegraphics[scale=0.6, width=130mm]{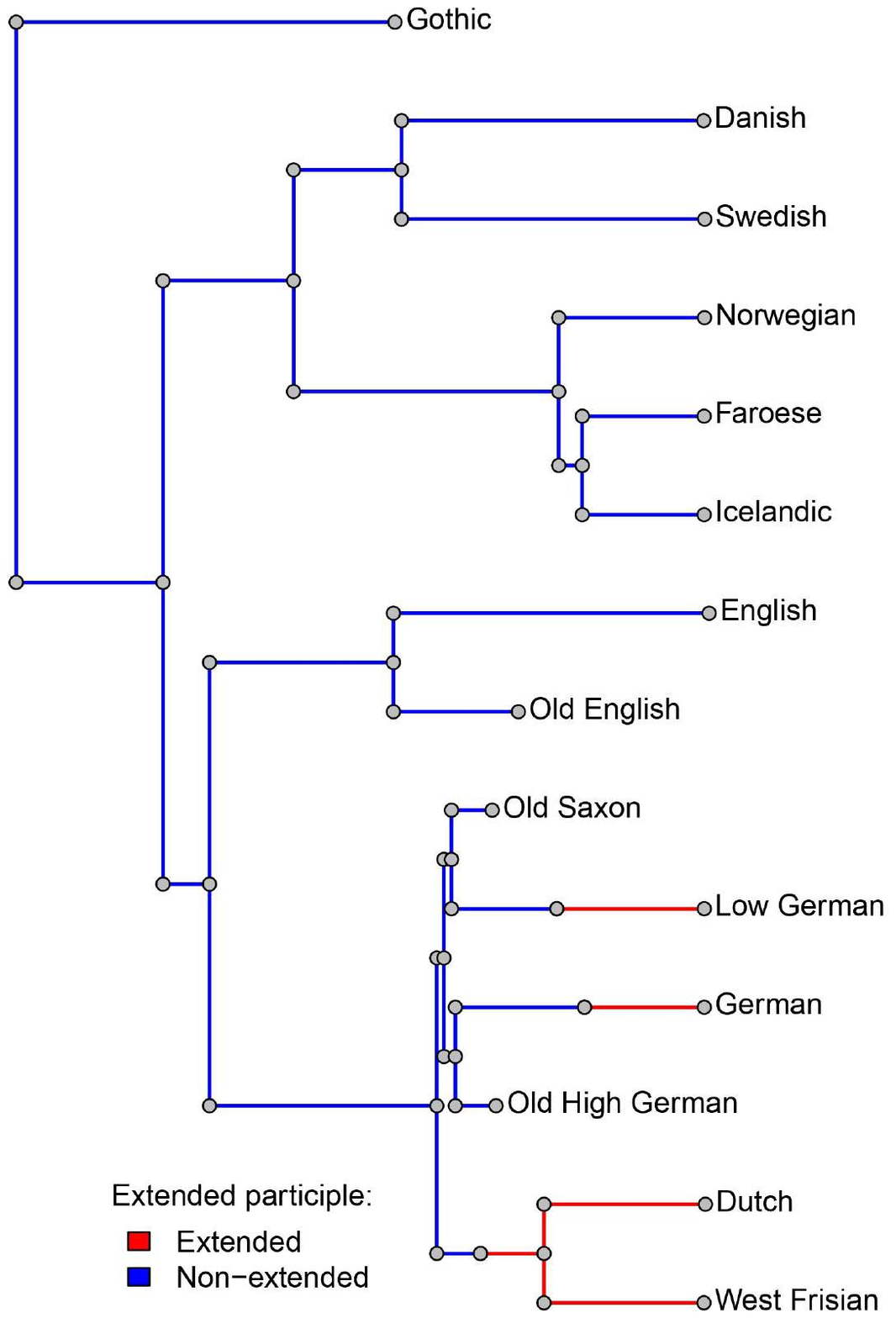}
    \caption{Stochastic character mapping for extended analytic perfect meaning character. The tree illustrated here is a Maximum Clade Credibility tree from \cite{chang2015}. 
    If color corresponds to ``extended'', it means that stochastic character mapping reconstructs probability of this state at this point in the branch being higher than 50\%.}
    \label{fig:transform_simmap}
\end{figure}

The data collected for this study apply only to the tips of the tree but not the internal nodes of the trees and not to certain points on branches. To solve this problem, we used stochastic character mapping, as implemented in \textsf{phytools} \citep{phytools}. Stochastic character mapping allows us to track the probability of change from a {\sc non-extended} participle to an {\sc extended} one at any moment along any branch. Note that we start with state {\sc N} at the root, as this is universally agreed to be an innovation in Germanic, as opposed to a retention \citep{seiler2022reichenbach}. In principle, one could use probabilities of the participle being extended in meaning to inject additional uncertainty in the model. However, this creates additional complications for the implementation, so we limited ourselves to a deterministic definition of regimes, i.e., the transitions occur either according to the {\sc E} regime, or {\sc N} regime. To do that, we track the probability of the {\sc E} being higher than 50\% along a branch and once the probability is higher than this threshold, the {\sc E} regime kicks in. This tree-painting procedure is illustrated in Figure \ref{fig:transform_simmap}.



The evolution of stem alternation patterns was modeled jointly for the 107 Germanic verbs in our sample. 
We make use of two model settings in our analyses. 
In the first of these, a flat, non-hierarchical model, all verbs evolve according to a single shared set of $41$ parameters that vary across the two TAM regimes described above. 
This rather restrictive assumption may be unrealistic, as different pressures may be active in different cognate classes; furthermore, 
inferred parameter values may be sensitive to skews in the data. 
As a more realistic alternative model, we employ a hierarchical model, which assumes that lexeme-level transition rates are drawn from hierarchical distributions corresponding to each transition type, analogous to random effects in a regression model. 
This allows us to account for idiosyncrasies in change patterns at the level of the individual verb while shedding light on global trends in the verbal system on the whole. 

The hierarchical model incorporates log-normally distributed priors over transition rates, with the hyperparameters $\mu$ (mean) and $\sigma$ (scale) governing the distributions for each transition type across verbs. These priors define normal distributions for log-transformed transition rates, allowing for variability across verbs while enforcing regularization to prevent overfitting. 
The lexeme-specific transition rates 
for each verb are modeled as deviations from the global 
expected value. 
This structure enforces partial pooling \citep{gelman2007data,mcelreath2020statistical} by shrinking verb-specific estimates toward the global trends defined by the two TAM regimes. Regime-level parameters (corresponding to {\sc N} and {\sc E}) are sampled independently. 

In summary, the model jointly analyzes the evolution of stem alternation patterns in 107 Germanic verbs using two approaches: a flat model with shared parameters and a hierarchical model that accounts for verb-level variability. The hierarchical model employs log-linear priors to enable partial pooling across verbs, regularizing verb-specific transition rates toward global trends defined by the TAM regimes. This approach balances the capturing of general patterns of change while accommodating individual differences among verbs.

We fit both models on 50 trees from the tree sample, each mapped according to the regimes representing the TAM states E(xtended) and N(ot extended). Fitting the model on multiple trees serves to incorporate phylogenetic uncertainty. Prior to fitting the model on real data, we validated our model on the basis of simulated data in order to ensure that our model fitting procedure is not prone to false positives.
We find that hierarchical model falsely detects an effect --- a decisive difference in stationary probability between the two regimes for a state --- in only 1 case, when using 99\% highest density interval, and in 2 cases when using 95\% highest density interval. That is, only 2 states in two separate trees are found to be decisively different across the two regimes out of 250 possible cases (50 trees $\times$ 5 states, excluding \textsc{d}).
In the aforementioned analyses, we do not make any prior assumptions about verb-level alternation patterns to be reconstructed to Proto-Germanic, 
but find that our model produces reconstructions that agree with expert ones for 89\% of the verbs in our data set (\ref{appendix.reconstruction}). 
As an additional validation, we run the hierarchical model on the MCC tree of the tree sample while constraining the ancestral pattern for each verb at the root of the tree to match expert reconstructions (\ref{appendix.constrained}). 


We use Pareto-smoothed importance sampling leave-one-out cross-validation (PSIS-LOO-CV, \citealt{vehtari2017practical}) estimation to compare the predictive performance of the hierarchical and non-hierarchical models, finding better performance for the hierarchical model and therefore justifying this model's additional complexity. 
The models are fitted using \textsf{RStan} \citep{Carpenteretal2017}. 
Further details of the model specification and inference along with a complete description of the simulation procedure can be found in the Appendix \ref{app:model}. 
Data and code are found at \url{https://gitlab.uzh.ch/alexandru.craevschi/germanic_strong_verbs}.


\begin{figure}
    \centering

        \begin{adjustbox}{max totalsize={\linewidth}{\linewidth},center}
        \includegraphics[width=\linewidth]{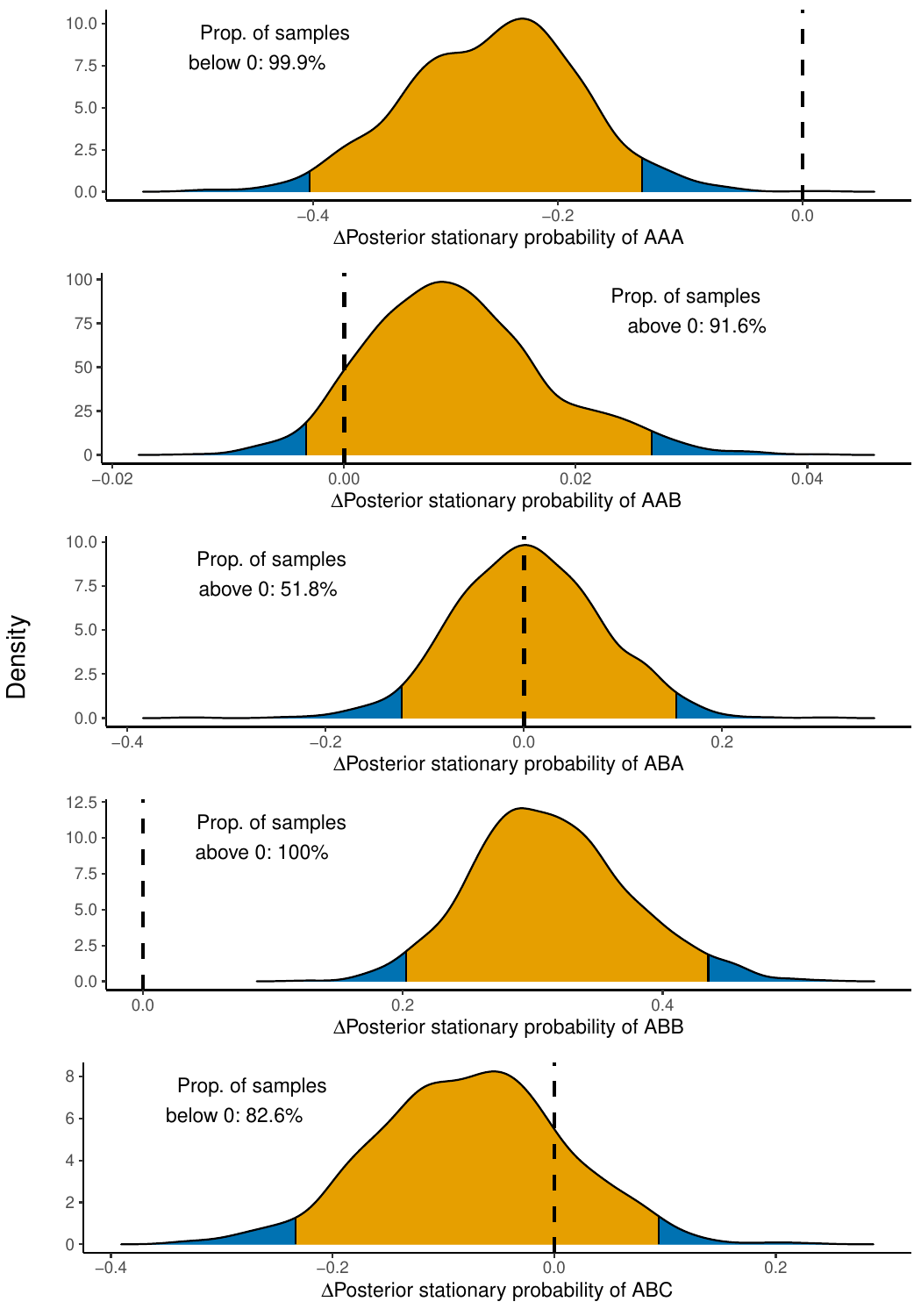}
        \end{adjustbox}
 
    \caption{Posterior distributions of differences in alternation pattern-level stationary probabilities (extended vs. non-extended).}
    \label{fig:stat.prob}
\end{figure}

\begin{figure}
    \centering
        
        \begin{adjustbox}{max totalsize={\linewidth}{\linewidth},center}
        \includegraphics[width=\linewidth]{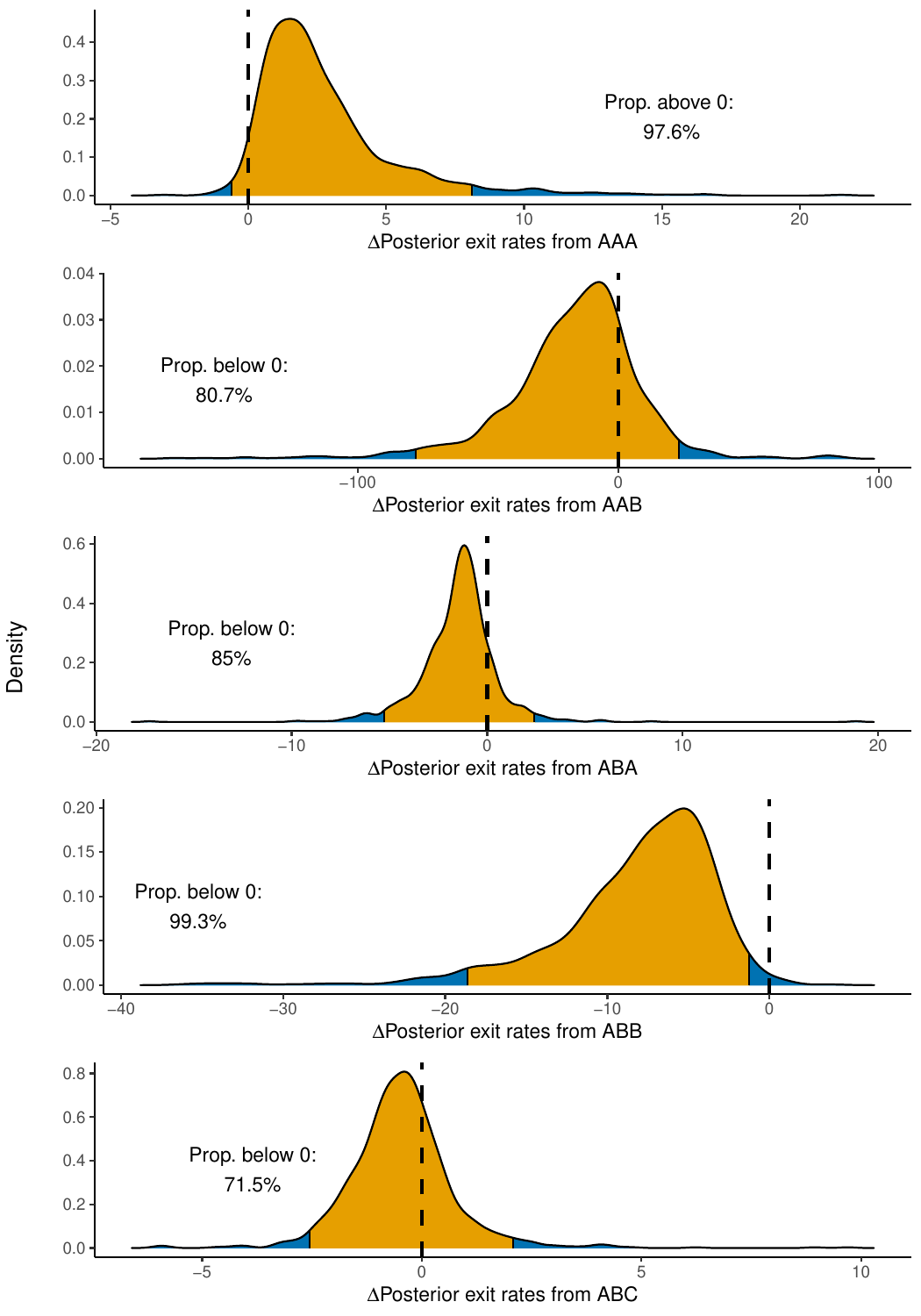}
        \end{adjustbox}

    \caption{Posterior distributions of differences in exit rates for different alternation patterns (extended vs. non-extended).}
    \label{fig:exit.rates}
\end{figure}

\begin{figure}
    \centering

        \begin{adjustbox}{max totalsize={\linewidth}{\linewidth},center}
        \includegraphics[width=\linewidth]{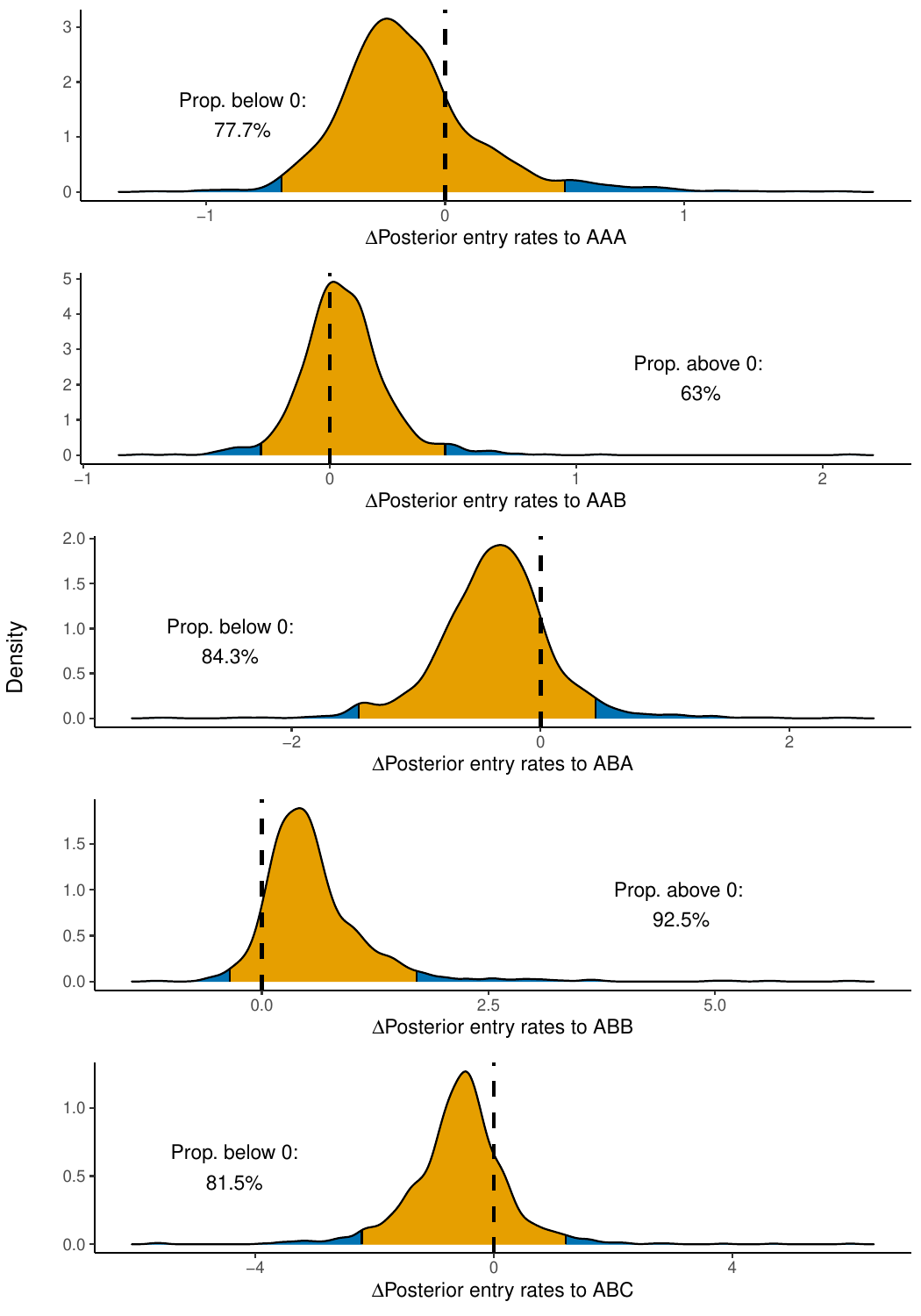}
        \end{adjustbox}

    \caption{Posterior distributions of differences in entry rates for different alternation patterns (extended vs. non-extended).}
    \label{fig:entry.rates}
\end{figure}





\section{Results}

The model fitting procedure yields 
samples from the posterior distributions of model parameters, in our case between-state CTM transition rates. 
We evaluate our main hypothesis of interest, namely whether there is a greater preference for the ABB pattern, using inferred posterior parameter values and quantities derived from these parameter values. 
For each posterior sample, we calculate the stationary distribution of the CTM process characterized by the sampled transition rates in order to construct a posterior distribution of stationary distributions, representing the long-term preferences for each alternation pattern in the extended and non-extended TAM regimes. 
We assess the degree to which the two TAM regimes differ in their propensities for different states by computing the pairwise differences in posterior samples between the extended and non-extended regimes for each alternation pattern's stationary probability. 
We calculated stationary probabilities by normalizing the right eigenvectors of the CTM rate matrix using R's \texttt{svd()} function. 
In line with relatively conservative Bayesian hypothesis assessment criteria \citep{kruschke2021bayesian}, we consider a difference to be decisive only if the 95\% highest posterior density interval (HDI) excludes zero, a null value representing no difference. 

Figure \ref{fig:stat.prob} displays posterior distributions of differences in alternation pattern-level stationary probabilities (extended vs. non-extended). 
As is clear, the long-term preference for ABB is decisively higher in the extended regime than in the non-extended one; the 95\% HDI of differences excludes zero, with 100\% of samples showing a difference greater than zero. 
Additionally, the long-term preference for AAA is decisively lower in the extended regime than in the non-extended one. 
This is at least partially 
a property of reallocation of mass on the probability simplex: increases in probability mass of one event coincide with decreases in probability mass of another event. 
Our results display a clear relative preference for ABB and relative dispreference for AAA under the extended regime. 
However, in principle, two scenarios could be at play here. 
It could be the case that extension leads to a higher preference for the ABB pattern, and the decreased preference for AAA on lineages with extension is simply an artifact of the increased preference for ABB. 
Conversely, it could be the case that lineages within the non-extended regime, several of which show various types of simplification of the verbal system (e.g., inflectional syncretism), some of which may be due to contact-induced language shift \citep{dahl2001circum,cole2022contact}, display a greater preference for the AAA pattern, and the decreased preference for ABB is an epiphenomenon of this. However, if the latter scenario were true, we would expect differences in preference for other alternation pattern types across the two regimes (viz., a decreased preference for all non-uniform alternation patterns in the non-extended regime), but we do not see this. 

To further clarify the dynamics of change that underlie the increased preference for the ABB pattern on extended branches, we inspect other properties of our inferred evolutionary parameters. 
For each state, we compute posterior exit and entry rates, representing the overall frequencies at which verb forms abandon a given alternation pattern for another or adopt a pattern 
(for details on how these quantities are calculated, see the appendix). 
These quantities help us assess whether the two regimes differ in terms of their attraction and repulsion dynamics 
to and from different alternation patterns. 

\begin{table}[]
\resizebox{\textwidth}{!}{%
\begin{tabular}{llllll}
\\
\multicolumn{1}{l|}{} &
  \multicolumn{1}{l|}{\textbf{AAA}} &
  \multicolumn{1}{l|}{\textbf{AAB}} &
  \multicolumn{1}{l|}{\textbf{ABA}} &
  \multicolumn{1}{l|}{\textbf{ABB}} &
  \textbf{ABC} \\
\multicolumn{1}{l|}{\textbf{$\Delta$Stationary probability}} &
  \multicolumn{1}{l|}{$\boldsymbol{[-0.403, -0.131]}$} &
  \multicolumn{1}{l|}{$[-0.003, 0.027]$} &
  \multicolumn{1}{l|}{$[-0.124, 0.154]$} &
  \multicolumn{1}{l|}{$\boldsymbol{[0.203, 0.435]}$} &
  {$[-0.233, 0.094]$} \\
\multicolumn{1}{l|}{\textbf{$\Delta$ Entry rate}} &
  \multicolumn{1}{l|}{$[-0.685, 0.501]$} &
  \multicolumn{1}{l|}{$[-0.279, 0.468]$} &
  \multicolumn{1}{l|}{$[-1.459, 0.444]$} &
  \multicolumn{1}{l|}{$[-0.352, 1.707]$} &
  {$[-2.214, 1.206]$} \\
\multicolumn{1}{l|}{\textbf{$\Delta$ Exit rate}} &
  \multicolumn{1}{l|}{$[-0.598, 8.113]$} &
  \multicolumn{1}{l|}{$[-77.781, 23.193]$} &
  \multicolumn{1}{l|}{$[-5.282, 2.407]$} &
  \multicolumn{1}{l|}{$\boldsymbol{[-18.63, -1.257]}$} &
  {$[-2.557, 2.076]$} \\
\\
\end{tabular}%
} 
\caption{\label{table:hdis} 95\% highest posterior density intervals of differences in alternation pattern-level stationary probabilities, entry and exit rates (extended vs. non-extended). Entries in bold face indicate intervals that do not overlap with 0 and display decisive evidence for a difference.}
\end{table} 

Figures \ref{fig:exit.rates} and \ref{fig:entry.rates} provide posterior distributions of differences in exit and entry rates for different alternation patterns (extended vs. non-extended). 
Table \ref{table:hdis} summarizes this information in a tabular format. 
According to our criteria, we find that there is no difference in entry rates across regimes for all alternation patterns, but the exit rate for the pattern ABB is decisively lower in the extended regime than in the non-extended one. 
This indicates that the chief mechanism responsible for the asymmetry in preferences across regimes is that verbs displaying the ABB pattern abandon this pattern less frequently in the extended regime than in the non-extended regime. 
Thus, the expansion of narrative past tense semantics to present perfect creates a state of affairs that facilitates the persistence and entrenchment of this semantically motivated stem alternation pattern; however, it is not necessarily the case that verbs that did not originally display the ABB pattern are more likely to be attracted to it in this regime. 

\section{Discussion}

Our results are twofold: first, we confirm a previously proposed association between the TAM semantics of the Germanic analytic perfect and the ABB stem alternation pattern using data from multiple Germanic languages and a phylogenetic model that explicitly characterizes the dynamics of linguistic evolution. 
Second, we shed light on the mechanisms that underlie this association, which could stem from a number of underlying factors. 
The evolutionary transition rates most likely to give rise to the data we observe suggest that the dynamic is largely one involving the retention of certain patterns under certain circumstances, rather than a scheme under which items are drawn to a beneficial pattern. This is the scenario envisioned by previous studies as being responsible for the association between extended narrative past tense and the ABB pattern (e.g., \citealt{desmet_vandevelde2019}, who frame this association in terms of the longer survival of ABB), but to our knowledge, our results are the first confirmation of this view capable of explicitly teasing apart a wide variety of mechanisms at play. 


This finding bears on debates in the literature on morphological change regarding irregularization versus the preservation of regularity, albeit indirectly. 
There is some debate as to whether irregularity is found in communicatively beneficial contexts (e.g., high-frequency items; \citealt{Blevinsetal2017}) due to changes that actively introduce irregularity \citep{Nuebling2000} or simply due to the maintenance of irregular patterns \citep{Gaeta2007} that result from orthogonal processes such as sound change \citep{sturtevant1947introduction}. 
While it is uncontroversial that Germanic languages undergo irregularization at times, e.g., German {\it R\"uckumlaut} \citep{fertig2017morphological}, the innovation of English past tense {\it dived} $\rightarrow$ {\it dove} \citep{newberry2017detecting}, etc., our results support the idea that the distribution of certain irregular, non-uniform paradigms (specifically ABB) overwhelmingly results from the conservation of a much older irregular pattern inherited from Proto-Indo-European whose origins are unclear but may be due to prosodic conditioning at an earlier stage of the proto-language \citep{lundquist2018morphology}. 



Some discussion is warranted regarding the causal interpretation of our results. 
Our results are compatible with a scenario under which present perfect constructions expand their meaning to be used for narrative past, which causes verbs to retain the ABB pattern with higher probability. 
At the same time, ours is not an explicit causal model: we have assumed a directional association from semantics to morphology, and have not allowed the causal chain between the two variables to fall out of the data. 
In our view, this assumption is justified; while it has been posited that certain morphological phenomena directly influence semantics and conceptualization 
\citep{Lucy1992,Acquaviva2004}, we know of no plausible, well-motivated claim that Germanic languages underwent extension of narrative past tense to perfect participles due to the higher frequencies of the ABB stem alternation pattern. 

Those familiar with phylogenetic comparative methods may question why we have assumed a directional association from semantics to morphology rather than employing certain standard tools for phylogenetic causal inference. 
The Discrete method and its variants \citep{Pagel1994,PagelMeade2006} explicitly model precedence relations in changes involving pairs of correlated features, which can be seen as an operationalization of causality. 
However, we are not working with a pair of two features, but rather with changes in TAM plus 107 cognate verbs. Extending the Discrete model to the analysis of the coevolution of 108 features would result in an intractable CTM state space that would need to take into account a number of value combinations 
that scales exponentially with the number of features. In contrast, our approach allows us to model individual verbs as evolving according to independently and identically distributed parameters with a factorizable likelihood, facilitating model comparison techniques such as PSIS-LOO-CV. 
It is worth noting that it is in theory possible to model the coevolution between verb-level patterns and TAM semantics independently for each verb in our sample, but this has the consequence of assuming that TAM semantics evolves independently across verbs rather than at the language level. While this approach is vaguely reminiscent of views of grammar which argue for an item-based approach to acquisition, usage and change \citep{goldberg2006constructions}, 
the latter assumption strikes us as less controversial and better motivated, particularly since we do not observe any synchronic intra-language variation in TAM semantics at the level of individual verbs (item-based approaches may be better suited to modeling other phenomena, such as changes in grammatical relations for individual verbs; \citealt{bardhdal2012reconstructing}). 

Another possible means of investigating causal relationships between coevolving variables is a recently developed method for continuous traits which infers whether changes in the value of one variable bring about changes in the ``optimal'' value of the other trait, i.e., the value to which it is expected to revert in the long term \citep{ringen2021novel,sheehan2023coevolution}. 
At the time of writing, this method has been used exclusively to explore relationships between pairs of variables; while it can be expanded to explore dependencies between more than two variables, it is not clear if it will scale well to more than a hundred variables, given computational considerations such as the construction of large covariance matrices. 
In theory, while this method should be extendable to non-ordinal categorical data, we have not yet explored the consequences that various choices vis \`a vis this model would have for analyzing the data we work with. 
However, as causal phylogenetic models grow more flexible, addressing questions of this sort will become more feasible.

\section{Conclusion}

This paper employed a phylogenetic model to explore the evolution of morphological patterns at the lexeme-level in the Germanic subgroup of Indo-European. 
We used a hierarchical model which allowed us to uncover an association between TAM semantics of the present perfect and a particular stem alternation pattern (ABB) while controlling for idiosyncratic behavior at the lexical level. While we did not explicitly include variables pertaining to usage-based properties like frequency, etc., it is likely that the structure of our model captures variability along these lines at least indirectly. 
As historical and synchronic resources are expanded and flexible phylogenetic models of this sort are further developed, it will be possible to investigate the role of a suite of factors underlying phenomena of this sort. 
The approach can be extended to address other questions regarding the evolution of patterns within lexical items and related phenomena. 

Our results confirm a hypothesis regarding the greater affinity for certain stem alternation patterns under certain TAM configurations. To our knowledge, we are the first to uncover this association using a phylogenetic model. This model can be seen simply as a means of demonstrating an association while accounting for Galton's problem \citep{Narroll1961}, i.e., the fact that patterns found in phylogenetically related taxa are not independent observations. However, the utility of such models does not end here: as we show, the transition rate parameters of phylogenetic models also provide information further clarifying the mechanisms underlying associations of this sort, and can directly engage with mechanistic explanations (e.g., cognitive, sociolinguistic) regarding how these patterns emerge and are maintained.


\section*{Funding information}

A.C.\ and S.B. were supported by Swiss National Science Foundation grant No.\ 207573, awarded to C.C. 
C.C.\ was supported by the NCCR Evolving Language (SNSF Agreement No. 51NF40\underline{\phantom{X}}180888). 


\section*{Abbreviations}
\begin{longtable}{p{.2\linewidth}p{.7\linewidth}}
    TAM & Tense/Aspect/Mood\\
    PG & Proto-Germanic\\
    MCC & Maximum clade credibility tree (way of summarizing a posterior tree sample)\\
    MRCA & Most recent common ancestor of two nodes in a tree\\
    E & Extended (narrative past meaning extended to present perfect)\\
    N & Non-extended\\
    PSIS-LOO-CV & Pareto-smoothed importance sampling leave-one-out cross-validation (method for comparing goodness of fit of multiple Bayesian models)\\
    CTM & Continuous-time Markov process (a stochastic process characterized by transition rates, i.e., frequencies between different states or feature values)\\
    HDI & Highest Density Posterior Interval (narrowest range of values containing a specified proportion of posterior samples, by default 95\%; conceptually, the range of posterior parameter values with highest support which may include or exclude parameter values compatible with certain hypotheses)\\
    AAA, ABB, etc. & See Section \ref{data}\\
\end{longtable}

\bibliographystyle{unified}
\bibliography{bibliography}

\bigskip

\appendix
\section{Model specification and details of inference}
\label{app:model}

\subsection{Hierarchical model}
\label{hierarchical-model}

\href{https://gitlab.uzh.ch/alexandru.craevschi/germanic_strong_verbs/-/blob/main/analysis/models_code/simmap_verb_hierarchy.stan#L105-112}{Transition rates} of our hierarchical model have the following distributions. 
The global death rate of verbs, which is state independent and does not vary across verbs or regimes $\delta \sim \text{LogNormal}(0,1)$.
A transition rate 
between states $i \neq j \in \{\text{AAA},\text{AAB},\text{ABA},\text{ABB},\text{ABC}\}$ 
for verb $v \in \{1,...,V\}$ 
in regime $r \in \{\text{\sc E(xtended)},\text{\sc N(on-extended)}\}$ 
is denoted by 
$\rho^{r}_{v,i,j} \sim \text{LogNormal}(\mu^r_{i,j},\sigma^r_{i,j})$, where $\mu^r_{i,j} \sim \text{Normal}(0,1), \sigma^r_{i,j} \sim \text{HalfNormal}(0,1)$. 

From each set of verb-specific rates (plus the death rate) we construct two 
\href{https://gitlab.uzh.ch/alexandru.craevschi/germanic_strong_verbs/-/blob/main/analysis/models_code/simmap_verb_hierarchy.stan?ref_type=heads#L2-33}{rate matrices} (one for each regime), 
$Q_v^r : r \in \{\text{\sc E},\text{\sc N}\}$; 
off-diagonal cells contain inter-state transition rates (as defined above); diagonal cells contain the negative sum of all other cells in the same row. 
The likelihoods of the patterns attested in cognates of verb $v \in \{1,...,V\}$, $P(X|Q,T) = \prod_{v=1}^V P(\boldsymbol{x_v}|Q_v,T)$, can be computed individually via the \href{https://gitlab.uzh.ch/alexandru.craevschi/germanic_strong_verbs/-/blob/main/analysis/models_code/simmap_verb_hierarchy.stan#L36-71}{pruning algorithm} \citep{felsenstein1981,felsenstein2004}, a dynamic program that exploits the conditional independence structure of directed acyclic graphs (such as phylogenies; cf.\ \citealt{pearl2022reverend}). 
In post-order traversal (i.e., moving from nodes directly ancestral to the tips of the tree toward the root), we compute the following partial likelihood for each node $n$ ancestral to each branch segment,
$$
\mathcal{L}^n_v(s) = \prod_{d \in D(n)} \sum_{t \in \mathbb{S}} P_{s \rightarrow t}(\ell_{n,d};Q_v^{r_{n,d}}) \mathcal{L}^d_v(t)
$$
where $D(n)$ denotes the descendants of node $n$; 
$\mathbb{S}$ the set of states; 
$P_{s \rightarrow t}(\ell_{n,d};Q_v^{r_{n,d}})$ the probability of transitioning from state $s$ to state $t$ over time interval $\ell_{n,d}$, the length of the branch segment between node $n$ and its descendant $d$, under the rates for verb $v$ associated with $r_{n,d}$, the regime of the branch segment between node $n$ and its descendant $d$ (via matrix exponentiation: $P(\ell_{n,d};Q_v^{r_{n,d}}) = \exp{(Q_v^{r_{n,d}} \ell_{n,d})}$); and $\mathcal{L}_v^d(t)$ the likelihood of state $t$ at node $d$. State likelihoods at tips of the tree will be either 1 (if they attest the state in question) or 0; for tips with missing data, all state likelihoods are set to 1. 
The likelihood for the entire tree at the root $r$ is $P(\boldsymbol{x_v}|Q_v,T) = \sum_{s \in \mathbb{S}} \pi(s)\mathcal{L}_v^r(s)$, where $\pi(s)$ is the prior probability of state $s$ at the root of the tree (which we set uniformly to 
$\frac{1}{\mathbb{S}}$). 

We draw posterior parameter samples using RStan \citep{Carpenteretal2017}. 
We construct an approximation to the posterior distribution of parameters that incorporates uncertainty over the space of phylogenies $\Psi$ using tree samples (denoted $\mathbb{T}$), $P(Q|X) \propto P(Q,X) = \allowbreak \int_\Psi P(X|Q,\Psi) \allowbreak P(Q) P(\Psi) dT \allowbreak \approx 
\frac{1}{|\mathbb{T}|} \sum_{t \in \mathbb{T}} P(X|Q,T) P(Q)$ 
by aggregating posterior samples across trees. 
For each tree in the 50-tree sample, we run \href{https://gitlab.uzh.ch/alexandru.craevschi/germanic_strong_verbs/-/blob/main/analysis/simmap_verb_hierarchy.R#L105-116}{10000 iterations} of the No U-Turn Sampler (NUTS) over 4 chains, discarding the first 8000 samples. We set the {\tt adapt\underline{\phantom{x}}delta} argument to $0.999$, which decreases the step size of the sampler and mitigates divergence across chains. 
We use the {\tt map\underline{\phantom{x}}rect()} function to parallelize the computation of verb-level likelihoods. 
We monitor convergence using the potential scale reduction factor \citep{gelman1992inference}, with values close to 1 indicating that all chains have mixed. 
We aggregate posterior samples together from individual trees in the sample using the function {\tt sflist2stanfit()}. 

Following rate inference, we use posterior expected rate values to address a number of hypotheses of interest. We construct \href{https://gitlab.uzh.ch/alexandru.craevschi/germanic_strong_verbs/-/blob/main/analyze_output/stationary_prob.R#L46-50}{posterior rate matrices} of expected transition rates under each regime, $Q_0^{r} : r \in \{\text{\sc e},\text{\sc n}\}$ from the expected rates $\exp{\left(\boldsymbol{\mu}\right)}$ and $\exp{\left(\delta\right)}$. We compute the \href{https://gitlab.uzh.ch/alexandru.craevschi/germanic_strong_verbs/-/blob/main/analyze_output/stationary_prob.R#L52-53}{posterior stationary distributions} $\boldsymbol{\pi}^r : r \in \{\text{\sc e},\text{\sc n}\}$ for each regime using R's {\tt svd()} function, normalizing the right eigenvalues. 
Entry and exit rates are calculated as follows (derived by Gerhard J\"ager {\it apud} \citealt[SI 8--10]{CarlingCathcart2021b}); 
below, $R(i \rightarrow j)$ denotes the transition rate from state $i$ to state $j$, and $\pi(j)$ denotes the stationary probability of state $j$:
\begin{itemize}
    \item \href{https://gitlab.uzh.ch/alexandru.craevschi/germanic_strong_verbs/-/blob/main/analyze_output/entry+exit_rates_new.R?ref_type=heads#L142-143}{Exit rate for state} $i$: $$\sum_{j \neq i} R(i \rightarrow j)$$
    \item \href{https://gitlab.uzh.ch/alexandru.craevschi/germanic_strong_verbs/-/blob/main/analyze_output/entry+exit_rates_new.R?ref_type=heads#L31-46}{Entry rate for state} $i$: $$\frac{\sum_{j \neq i} \pi(j) R(j \rightarrow i)}{\sum_{j \neq i} \pi(j)}$$
\end{itemize}
We construct posterior distributions of differences in stationary probabilities, entry, and exit rates between the extended and non-extended regimes by taking the differences of these parameters across samples. 
We compute HDIs with the R package {\tt HDInterval} \citep{HDI}. 

\subsection{Non-hierarchical model}

Our non-hierarchical setting assumes that rates of change are \href{https://gitlab.uzh.ch/alexandru.craevschi/germanic_strong_verbs/-/blob/main/analysis/models_code/simmap_no-hierarchy.stan?ref_type=heads#L104-105}{invariant across cognate verbs}, and that each verb $v \in \{1,...,V\}$ evolves according to the rates $\exp \left( \boldsymbol{\mu} \right)$ and $\exp \left( \delta \right)$. 
Details of inference are as above. 

\subsection{Model comparison}
We carry out model comparison between the \href{https://gitlab.uzh.ch/alexandru.craevschi/germanic_strong_verbs/-/blob/main/model_comparison/compare_models.R}{hierarchical and flat models
using PSIS-LOO-CV} \citep{vehtari2017practical}. 
The hierarchical model greatly outperforms the flat model; between-model differences exceed $\pm$2 standard errors of the differences ($\text{ELPD}_{\text{\sc hier}} - \text{ELPD}_{\text{\sc flat}} = 57.90, \text{SE}(\Delta \text{ELPD}) = 11.87$).





\subsection{Ancestral state reconstruction}
\label{appendix.reconstruction}

We use \href{https://gitlab.uzh.ch/alexandru.craevschi/germanic_strong_verbs/-/blob/main/anc_rec/anc_rec.R#L106-115}{fitted rate parameters from the hierarchical model} to reconstruct ancestral state distributions for each verb in our data set to the root of the tree, representing Proto-Germanic. 
The probability of state $s$ at the root of the tree is proportional to the quantity $\pi(s)\mathcal{L}_v^r(s)$, \href{https://gitlab.uzh.ch/alexandru.craevschi/germanic_strong_verbs/-/blob/main/anc_rec/anc_rec.R#L119}{normalized so that all probabilities sum to $1$} \citep{yang2014molecular}. 
We carry out this procedure for each posterior rate sample, sampling a state from the categorical distribution parameterized by the resulting distribution of states at the root, and compute the posterior probability of each sampled state. 
For each verb, we \href{https://gitlab.uzh.ch/alexandru.craevschi/germanic_strong_verbs/-/blob/main/anc_rec/eval_rec.R?ref_type=heads}{compare the state with highest probability to the state reconstructed to Proto-Germanic by experts} (we rely primarily on the paradigmatic information found in Wiktionary, which is taken primarily from \citealt{ringe2008proto}). 
According to this procedure, our model reconstructs ancestral states with 89\% accuracy, correctly reconstructing 95 out of 107 verbal patterns for Proto-Germanic.


\subsection{Ancestry-constrained model}
\label{appendix.constrained}

As an additional validation of our results, we ran an ancestry-constrained version of the hierarchical model, intended to assess whether we obtain similar results when we enforce the state reconstructed at the root of the tree to match expert reconstructions arrived at via the comparative method. 
We repeat the procedure described for the hierarchical model (due to the computationally intensive nature of model fitting, we limit ourselves to running the model on the \href{https://gitlab.uzh.ch/alexandru.craevschi/germanic_strong_verbs/-/blob/main/analysis_results/tam_simmap_root/summ_simmap_mcc.rds?ref_type=heads}{MCC tree of the tree sample}), with one key difference: for each verbal cognate set, we attach an \href{https://gitlab.uzh.ch/alexandru.craevschi/germanic_strong_verbs/-/blob/main/analysis/tam_simmap_root.R?ref_type=heads#L20-21}{``adventitious'' branch of very short length to the root of the tree}, setting the state of the descendant node to the state reconstructed to Proto-Germanic by experts (we rely primarily on the paradigmatic information found in Wiktionary). 
Under this setting, we find the same key result as for the hierarchical model, i.e., that there is a higher stationary probability of ABB and lower stationary probability of AAA in the extended regime than in the non-extended regime. 

\subsection{Simulation-based validation study}

We conducted a simulation study designed to assess the extent to which our hierarchical model-fitting procedure yields false positives and validate whether our result reported in this paper can be trusted. 
Specifically, we \href{https://gitlab.uzh.ch/alexandru.craevschi/germanic_strong_verbs/-/blob/main/simulation/sim_history.R?ref_type=heads#L37-78}{simulated data under a non-hierarchical model} with no differences in behavior across TAM regimes (extended vs.\ non-extended), and then \href{https://gitlab.uzh.ch/alexandru.craevschi/germanic_strong_verbs/-/blob/main/simulation/sim_history.R?ref_type=heads#L81-116}{fitted the hierarchical model on this synthetic data} to examine whether it erroneously detects differences in stationary probabilities between the two regimes. 
If the hierarchical model infers systematic differences between regimes where none exist, this would indicate an elevated false positive rate and potential model overfitting. 

We generated synthetic character matrices. We simulated character evolution for \href{https://gitlab.uzh.ch/alexandru.craevschi/germanic_strong_verbs/-/blob/main/simulation/sim_history.R?ref_type=heads#L24-25}{$100$ hypothetical verbs across $50$ phylogenetic trees}. The trees were the same as the ones used in the main analysis. The evolutionary process followed a non-hierarchical model where each verb shared a single set of transition rates across states, without any variation across regimes. This ensures that any inferred differences between regimes in the hierarchical analysis stem from model artifacts rather than genuine underlying patterns.

Each verb's transition rates were drawn from a log-normal hyper-distribution with parameters $\mu = 0.5$ and $\sigma = 0.1$. 
Death rates (i.e., transitions to the absorbing state \textsc{d}) were fixed at an arbitrarily small rate of $0.05$. 
We initialized root states such that all verbs start in the ABC state, reflecting the predominant reconstructed Proto-Germanic pattern. Evolutionary trajectories for each verb were simulated along the tree using the {\tt sim.history()} function from the {\tt phytools} \citep{phytools} package, which implements continuous-time Markov process simulations.



Details of inference are as described above (\ref{hierarchical-model}), except that we ran NUTS 
for \href{https://gitlab.uzh.ch/alexandru.craevschi/germanic_strong_verbs/-/blob/main/simulation/sim_history.R?ref_type=heads#L107-116}{3000 iterations across 4 chains, discarding the first 1500 samples.} 
We then compared inferred stationary probabilities between the extended and non-extended regimes (the same procedure carried out for our main analyses and displayed in Figure \ref{fig:stat.prob}). Since the data were generated without regime-specific effects, any decisive difference inferred by the hierarchical model would indicate a false positive.

As mentioned in the main text, we obtained a very low false positive rate. If we use a 99\% HDI to assess whether zero is included in the posterior credible interval, we detected only one state in one tree when the difference was deemed decisive according to our testing procedure (yielding a false positive rate of $0.01$). For a 95\% HDI, there were two states in two separate trees that were decisively different in their stationary probability ($0.02$). Finally, under an 89\% HDI, we had a false positive rate of $0.06$. For the paper, we report results with a 95\% HDI.








\end{document}